\documentclass{article}

\usepackage{amsmath,amsfonts,bm,amsthm,amssymb}

\def\eqref#1{equation~\ref{#1}}
\def\Eqref#1{Eq.~(\ref{#1})}

\def\Algref#1{Algorithm~\ref{#1}}

\def\1{\bm{1}}

\def\rvepsilon{{\mathbf{\epsilon}}}

\def\rvkappa{{\mathbf{\kappa}}}

\def\vmu{{\bm{\mu}}}

\def\va{{\bm{a}}}

\def\vq{{\bm{q}}}

\def\vs{{\bm{s}}}

\def\vx{{\bm{x}}}
\def\vy{{\bm{y}}}

\def\mI{{\bm{I}}}

\def\mSigma{{\bm{\Sigma}}}

\DeclareMathAlphabet{\mathsfit}{\encodingdefault}{\sfdefault}{m}{sl}
\SetMathAlphabet{\mathsfit}{bold}{\encodingdefault}{\sfdefault}{bx}{n}

\def\gA{{\mathcal{A}}}
\def\gB{{\mathcal{B}}}

\def\gD{{\mathcal{D}}}

\def\gJ{{\mathcal{J}}}

\def\gL{{\mathcal{L}}}

\def\gN{{\mathcal{N}}}
\def\gO{{\mathcal{O}}}
\def\gP{{\mathcal{P}}}

\def\gR{{\mathcal{R}}}
\def\gS{{\mathcal{S}}}
\def\gT{{\mathcal{T}}}

\def\sR{{\mathbb{R}}}

\def\gA{{\mathcal{A}}}
\def\gB{{\mathcal{B}}}

\def\gD{{\mathcal{D}}}

\def\gJ{{\mathcal{J}}}

\def\gL{{\mathcal{L}}}

\def\gN{{\mathcal{N}}}
\def\gO{{\mathcal{O}}}
\def\gP{{\mathcal{P}}}

\def\gR{{\mathcal{R}}}
\def\gS{{\mathcal{S}}}
\def\gT{{\mathcal{T}}}

\newcommand{\E}{\mathbb{E}}

\newcommand{\R}{\mathbb{R}}

\newcommand{\KL}{D_{\mathrm{KL}}}

\def\eqref#1{equation~\ref{#1}}
\def\Eqref#1{Eq.~(\ref{#1})}

\def\Algref#1{Algorithm~\ref{#1}}

\newcommand{\cbr}[1]{\left\{#1\right\}}
\newcommand{\sbr}[1]{\left[#1\right]}

\DeclareMathOperator*{\argmax}{arg\,max}
\DeclareMathOperator*{\argmin}{arg\,min}

\usepackage{microtype}
\usepackage{graphicx}
\usepackage{subfigure}
\usepackage{booktabs} 

\usepackage{tabularray} 
\UseTblrLibrary{booktabs}
\usepackage{algorithm}
\usepackage{algorithmic}

\usepackage{url}
\usepackage{comment}
\usepackage{tabularx}

\usepackage{hyperref}

\usepackage[accepted]{icml2024}

\usepackage{amsmath}
\usepackage{amssymb}
\usepackage{mathtools}
\usepackage{amsthm}

\usepackage[capitalize,noabbrev]{cleveref}

\theoremstyle{plain}
\newtheorem{theorem}{Theorem}[section]
\newtheorem{proposition}[theorem]{Proposition}
\newtheorem{lemma}[theorem]{Lemma}
\newtheorem{corollary}[theorem]{Corollary}
\theoremstyle{definition}
\newtheorem{definition}[theorem]{Definition}
\newtheorem{assumption}[theorem]{Assumption}
\theoremstyle{remark}
\newtheorem{remark}[theorem]{Remark}

\usepackage[textsize=tiny]{todonotes}
\allowdisplaybreaks[4]

\icmltitlerunning{DiffCPS: Diffusion-based Constrained Policy Search 
for Offline Reinforcement Learning}

\begin{document}

\twocolumn[
\icmltitle{DiffCPS: Diffusion-based Constrained Policy Search for Offline Reinforcement Learning}



\icmlsetsymbol{equal}{*}

\begin{icmlauthorlist}
\icmlauthor{Longxiang He}{yyy}
\icmlauthor{Li Shen}{comp}
\icmlauthor{Linrui Zhang}{yyy}
\icmlauthor{Junbo Tan}{yyy}
\icmlauthor{Xueqian Wang}{yyy}
\end{icmlauthorlist}

\icmlaffiliation{yyy}{Center for Artificial Intelligence and Robotics,Tsinghua University
, Shenzhen, China}
\icmlaffiliation{comp}{JD Explore Academy, Beijing, China}

\icmlcorrespondingauthor{Li Shen}{mathshenli@gmail.com}
\icmlcorrespondingauthor{Xueqian Wang}{wang.xq@sz.tsinghua.edu.cn}

\icmlkeywords{Machine Learning, ICML}

\vskip 0.3in
]



\printAffiliationsAndNotice{}  

\begin{abstract}
Constrained policy search (CPS) is a fundamental problem in offline reinforcement learning, which is generally solved by advantage weighted regression (AWR). However, previous methods may still encounter out-of-distribution actions due to the limited expressivity of Gaussian-based policies.  On the other hand, directly applying the state-of-the-art models with distribution expression capabilities (i.e., diffusion models) in the AWR framework is intractable since AWR requires exact policy probability densities, which is intractable in diffusion models. In this paper, we propose a novel approach, \emph{Diffusion-based Constrained Policy Search} (dubbed DiffCPS), which tackles the diffusion-based constrained policy search with the primal-dual method.  The theoretical analysis reveals that strong duality holds for diffusion-based CPS problems, and upon introducing parameter approximation, an approximated solution can be obtained after $\gO(1/\epsilon)$ number of dual iterations, where $\epsilon$ denotes the representation ability of the parametrized policy. Extensive experimental results based on the D4RL benchmark demonstrate the efficacy of our approach. We empirically show that DiffCPS achieves better or at least competitive performance compared to traditional AWR-based baselines as well as recent diffusion-based offline RL methods. 
\end{abstract}

\section{Introduction}
Offline Reinforcement Learning (offline RL) aims to seek an optimal policy without environmental interactions~\citep{fujimoto2019,levine2020}. This is compelling for having the potential to transform large-scale datasets into powerful decision-making tools and avoids costly and risky online data collection, which offers significant application prospects in fields such as healthcare~\citep{nie2021learning,tseng2017deep} and autopilot~\citep{yurtsever2020survey,rhinehart2018deep}. 

Notwithstanding its promise, applying contemporary off-policy RL algorithms~\citep{lillicrap2015,fujimoto2018,haarnoja2018,haarnoja2018a} directly into the offline context presents challenges due to distribution shift \citep{fujimoto2019,levine2020}.
Previous methods to mitigate this issue under the model-free offline RL setting generally fall into three categories: 1) value function-based approaches, which implement pessimistic value estimation by assigning low values to out-of-distribution actions~\citep{kumar2020,fujimoto2019}, 2) sequential modeling approaches, which casts offline RL as a sequence generation task with return guidance~\citep{chen2021,janner2022,liang2023,ajay2022}, and 3) constrained policy search (CPS) approaches, which regularizes the discrepancy between the learned policy and behavior policy~\citep{peters2010,peng2019,nair2020}. We focus on the CPS-based offline RL due to its convergence guarantee and outstanding performance in a wide range of tasks.

Prior solutions for CPS~\citep{peters2010,peng2019,nair2020} primarily train a parameterized unimodal Gaussian policy through weighted regression. However, recent works~\citep{chen2022,hansen-estruch2023,shafiullah2022behavior} show such unimodal Gaussian models in weighted regression will impair the policy performance due to limited distributional expressivity. For example, if we fit a multi-modal distribution with an unimodal Gaussian distribution, it will unavoidably result in covering the low-density area between peaks. Intuitively, we can choose a more expressive model to eliminate this issue. Nevertheless,  \citet{ghasemipour2020emaq} shows that VAEs~\citep{kingma2013} in BCQ~\citep{fujimoto2019} do not align well with the behavior dataset, which will introduce the biases in generated actions. \citet{chen2022} utilizes the diffusion probabilistic model~\citep{sohl-dickstein2015,ho2020,song2019} to generate actions and select the action through the action evaluation model under the well-studied AWR framework. However, AWR requires an exact probability density of behavior policy, which is still intractable for the diffusion-based one. Alternatively, they leverage Monte Carlo sampling to approximate the probability density of behavior policy, which inevitably introduces estimation biases and increases the cost of inference. According to motivating examples in Section \ref{toy}, we visually demonstrate how these issues are pronounced even on a simple bandit task. 

To solve the above issues, we present \textbf{Diffusion-based Constrained Policy Search (DiffCPS)} which incorporates the diffusion model to solve the limited expressivity problem of unimodal Gaussian policy. To avoid the exact probability density required by AWR, we solve the diffusion-based CPS problem with the primal-dual method. Thereby, DiffCPS solves the limited policy expressivity problem while avoiding the intractable density calculation brought by AWR.
Interestingly, we find that when we use diffusion-based policies, we can eliminate the policy distribution constraint in the CPS through the action distribution of the diffusion model. Then under some mild assumptions, we show that strong duality holds for the diffusion-based CPS problem, which makes it easily solvable using primal-dual methods. 
We also study the effect of using a finite parametrization for the policies and show that the price to pay in terms of the duality gap depends on the representation ability of the parametrization. And the expressiveness of the model is precisely the greatest strength of diffusion models. 

In summary, our main contributions are three-fold: 1)We explicitly eliminate the policy distribution constraint in the CPS problem. This allows us to achieve strong duality under some assumptions. 2)We present DiffCPS, which tackles the diffusion-based constrained policy search without resorting to AWR. Our theoretical analysis shows that our method can achieve near-optimal results by solving the primal-dual problem with a parametric policy. 3)Our experimental results illustrate superior or competitive performance compared to existing offline RL methods in D4RL tasks, which also corroborates the results of our theoretical analysis. Even compared to other diffusion-based methods, DiffCPS achieves state-of-the-art (SOTA) performance in D4RL MuJoCo locomotion and AntMaze average score, which substantiates the effectiveness of our method.


   \begin{figure*}[t]
        \centering
        \includegraphics[width=.9\textwidth]{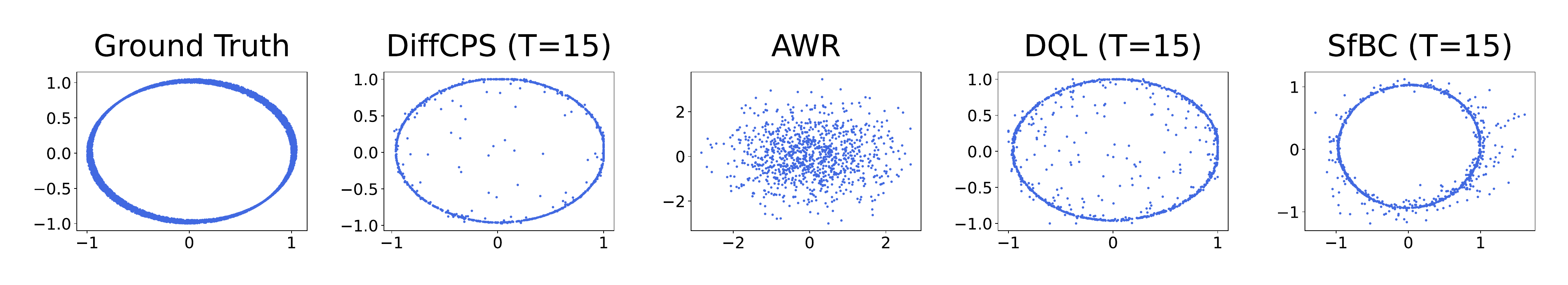} \\
        \includegraphics[width=.9\textwidth]{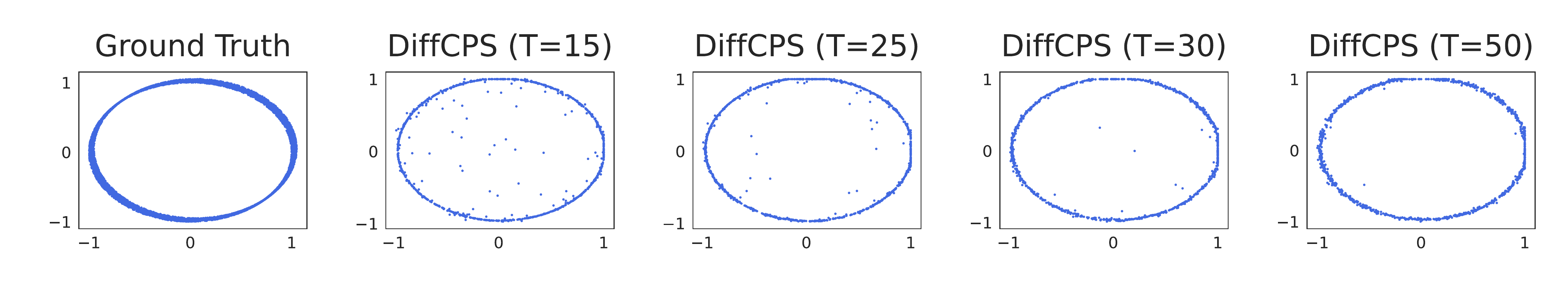}
        \vspace{-4.mm}
        \caption{Toy offline experiment on a simple bandit task. We test the performance of AWR and other diffusion-based offline RL algorithms (DQL~\citep{wang2022b} and SfBC~\citep{chen2022}). The first row displays the actions taken by the trained policy where $T$ denotes diffusion steps. We note that the AWR fails to capture the multi-modal actions in the offline dataset due to the limited policy expressivity of unimodal Gaussian. The second row shows the effect of different diffusion steps $T$.
        }
        \label{fig:toy_exp}
        \vspace{-2.mm}
    \end{figure*}

\section{Preliminaries}
\label{preliminary}

\textbf{Notation.} In this paper, we use $\mu_\theta(\va\vert\vs)$ or $\mu$ to denote the learned policy with parameter $\theta$ and $\pi_b$ to denote behavior policy that generated the offline data. We use superscripts $i$ to denote the diffusion time step and subscripts $t$ to denote RL trajectory timestep. For instance, $a_t^i$ denotes the  $t$-$\text{th}$ action in $i$-$\text{th}$ diffusion time step.

\subsection{Constrained Policy Search in Offline RL}


Consider a Markov decision process (MDP): $M = \{\gS, \gA, P, R, \gamma, d_0\}$, with state space $S$, action space $\gA$, environment dynamics $\gP(\vs' \vert \vs, \va): S \times  S \times \gA \rightarrow [0,1]$, reward function $R: S \times \gA \rightarrow \sR$, discount factor $\gamma \in [0, 1)$, policy  $\mu \in \gP(\gS)$, where $\gP(\gS)$ is the space of probability measures on~$(\gA, \gB(\gA))$ parametrized by elements of~$\gS$, where~$\gB(\gA)$ are the Borel sets of~$\gA$, and initial state distribution $d_0$. The action-value or Q-value of policy $\mu$ is defined as $Q^\mu(\vs_t, \va_t) = \E_{\va_{t+1}, \va_{t+2}, ... \sim \mu}\sbr{\sum_{j=0}^{\infty}\gamma^j r(\vs_{t+j}, \va_{t+j})}$. Our goal is  to get a policy to maximize  the cumulative discounted reward 
$J(\theta) = \int_\mathcal{S} d_0(\vs) Q^\mu(\vs, \va) d\vs$, where $\rho^\mu(\vs)=(1-\gamma)\sum_{t=0}^\infty\gamma^t p_\mu(\vs_t =\vs)$   is the normalized discounted state visitation frequencies induced by the policy $\mu$ and $p_\mu(\vs_t =\vs)$ is the likelihood of the policy being in state $\vs$ after following $\mu$ for $t$ timesteps~\citep{sutton2018,peng2019}. 

In offline setting~\citep{fujimoto2019}, environmental interaction is not allowed, and a static dataset $\gD \triangleq \cbr{(\gS, \gA, R, \gS', \text{done})}$ is used to learn a policy.
To avoid out-of-distribution actions, we restrict the learned policy $\mu$ to be not far away from the behavior policy $\pi_b$ by the KL divergence constraint. Prior works~\citep{peters2010,peng2019} formulate the above offline RL as a constrained policy search (CPS) problem with the following form:
\begin{align}\label{origin_cps}
        \mu^{*} = \argmax_\mu \ &J(\mu) = \argmax_\mu \int_\mathcal{S} d_0(\vs) \int_\mathcal{A}Q^\mu(\vs, \va) d\va d\vs  \nonumber\\
         s.t.  \quad & \KL(\pi_b(\cdot\vert\vs)\|\mu(\cdot\vert\vs)) \leq \rvepsilon,\quad \forall \vs\\  
         & \int_\va \mu(\va\vert\vs)d\va =1,\quad \forall \vs, \nonumber
\end{align}
Previous works~\citep{peters2010,peng2019,nair2020} solve \Eqref{origin_cps} through KKT conditions and get the optimal policy $\pi^*$ as:
\begin{align}
    \pi^*(\va\vert\vs) &= \frac{1}{Z(\vs)} \ \pi_b(\va\vert\vs) \ \mathrm{exp}\left(\alpha Q_\phi(\vs, \va) \right),
\label{Eq:pi_optimal}
\end{align}
where $Z(\vs)$ is the partition function. Intuitively we can use \Eqref{Eq:pi_optimal} to optimize policy $\pi$. However, the behavior policy may be very diverse and hard to model. To avoid modeling the behavior policy, prior works~\citep{peng2019,wang2020,chen2020} optimize $\pi^{*}$ through a parameterized policy $\pi_\theta$, known as AWR:
\begin{align}
    & \mathop{\mathrm{arg \ min}}_{\theta} \mathbb{E}_{\vs \sim \mathcal{D}^\mu} \left[ \KL \left(\pi^*(\cdot  | \vs) \middle|\middle| \pi_\theta(\cdot  | \vs)\right) \right]  \label{Eq:wr} \\
    = & \mathop{\mathrm{arg \ max}}_{\theta} \mathbb{E}_{(\vs, \va) \sim \mathcal{D}^\mu} \left[ \frac{1}{Z(\vs)} \mathrm{log} \ \pi_\theta(\va | \vs) \ \mathrm{exp}\left(\alpha Q_\phi(\vs, \va) \right) \right].\nonumber
\end{align}
where  $\mathrm{exp}(\alpha Q_\phi(\vs, \va))$ being the regression weights. However, AWR requires the exact probability density of policy, which restricts the use of generative models like diffusion models. In this paper, we directly utilize the diffusion-based policy to address Eq.~(\ref{origin_cps}). Therefore, our method not only avoids the need for explicit probability densities in AWR but also solves the limited policy expressivity problem.

\subsection{Diffusion Model}

\textbf{Diffusion Probabilistic Model (DPM).} 
Diffusion models~\citep{sohl-dickstein2015,ho2020,song2019} are composed of two processes: the forward diffusion process and the reverse process. In the forward diffusion process, we gradually add Gaussian noise to the data $\vx_0 \sim q(\vx_0)$ in $T$ steps. The step sizes are controlled by a variance schedule $\beta_i$:
\begin{equation}
\begin{aligned}
                &q(\vx_{1:T}  \,|\,  \vx_0) :=  \textstyle \prod_{i=1}^T q(\vx_i  \,|\,  \vx_{i-1}), \\
                &q(\vx_i  \,|\,  \vx_{i-1}) := \gN (\vx_i; \sqrt{1 - \beta_i} \vx_{i-1}, \beta_i %
        \mI).
\end{aligned}
\end{equation}

In the reverse process, we can recreate the true sample $\vx_0$ through $p(\vx^{i-1}\vert\vx^{i})$:
\begin{equation}
    \begin{aligned}
            p(\vx) & = \int p(\vx^{0:T})d\vx^{1:T}\\
            &=\int\gN (\vx^T; \mathbf{0},\mI)\prod_{i=1}^Tp(\vx^{i-1}\vert\vx^{i})d\vx^{1:T}.
    \end{aligned}
    \label{reverse}
\end{equation}
The training objective is to maximize the ELBO of $\E_{\vq_{x_0}}\sbr{\log p(\vx_0)}$. Following DDPM~\citep{ho2020}, we  use the simplified surrogate loss $\gL_d(\theta) = \E_{i \sim \sbr{1,T}, \rvepsilon \sim \gN(\mathbf{0}, \mI), \vx_0 \sim q} \sbr{|| \rvepsilon - \rvepsilon_\theta(\vx_i, i) ||^2}$ to approximate the ELBO. After training, sampling from the diffusion model is equivalent to running the reverse process. 

\textbf{Conditional DPM.} 
There are two kinds of conditioning methods: classifier-guided~\citep{dhariwal2021} and classifier-free~\citep{ho2022}. The former requires training a classifier on noisy data $\vx_i$ and using gradients $\nabla_\vx\log f_\phi(\vy\vert\vx_i)$ to guide the diffusion sample toward the conditioning information $\vy$. The latter does not train an independent $f_\phi$ but combines a conditional noise model $\epsilon_\theta(\vx_i, i,\vs)$ and an unconditional model $\epsilon_\theta(\vx_i, i)$ for the noise. The perturbed noise $w\epsilon_\theta(\vx_i, i)+(w+1)\epsilon_\theta(\vx_i,i,\vs)$ is used to later generate samples. However \citet{pearce2023imitating} shows this combination will degrade the policy performance in offline RL. Following~\citet{pearce2023imitating,wang2022b} we solely employ a conditional noise model $\epsilon_\theta(\vx_i,i,\vs)$ to construct our noise model ($w=0$).

\section{Methodology}

In this section, we detail the design of our Diffusion-based Constrained Policy Search (DiffCPS). First, we show that the limited policy expressivity in the AWR will degrade the policy performance through the bandit problem (Section \ref{toy}). Second, we formulate CPS via diffusion-based policy and get our diffusion-based CPS problem (Section \ref{DIffCPS-model}). Then, we solve DiffCPS with primal-dual method and show the price to pay in terms of the duality gap depending on the representation ability of the parametrization (Sectioon \ref{pd-solution-diffcps}). (All proofs are placed in Appendix~\ref{proof} due to limited space.)

\subsection{Is policy expressivity all you need for AWR?}
\label{toy}

Before showing our method, we first present that the limited policy expressivity in previous Advantage Weighted Regression (AWR) methods~\citep{peters2010,peng2019,nair2020} may degrade the performance through a sample 2-D bandit experiment with real-valued actions. Our offline dataset is constructed by a unit circle with noise (The first panel of Figure \ref{fig:toy_exp}). Data on the noisy circle have a positive reward of $1$. Note that this offline dataset exhibits strong multi-modality since there are many actions (points on the noisy circle) that can achieve the same reward of $1$ if a state is given. However, if we use unimodal Gaussian to represent the policy, although the points on the noisy circle can all receive the same reward of $1$, the actions taken by AWR will move closer to the center of the noisy circle (AWR in Figure \ref{fig:toy_exp}), due to the incorrect unimodal assumption. Experiment results in Figure \ref{fig:toy_exp} illustrate that AWR performs poorly in the bandit experiments compared to other diffusion-based methods. We also notice that compared to other diffusion-based methods, actions generated by SfBC include many points that differ significantly from those in the dataset, when $T=15$. This is due to the sampling error introduced by incorporating diffusion into AWR.

Therefore, we conclude that policy expressivity is important in offline RL since most offline RL datasets are collected from multiple behavior policies or human experts, which exhibit strong multi-modality. To better model behavior policies, we need more expressive generative models to model the policy distribution, rather than using unimodal Gaussians. Furthermore, we also need to avoid the sampling errors caused by using diffusion in AWR, which is the motivation behind our algorithm.

\subsection{Diffusion-Based Constrained Policy Search}\label{DIffCPS-model}

In this subsection, we focus on deriving our diffusion-based CPS problem \Eqref{cps}. Firstly, we present the form of constrained policy search~\citep{peng2019}:
\begin{align}
\label{obj}
        \mu^{*} \!=\! \argmax_\mu \ &J(\mu) \!=\! \argmax_\mu \int_\mathcal{S} d_0(\vs) \int_\mathcal{A}Q^\mu(\vs, \va) d\va d\vs\\
        \label{kl}
         s.t.  & E_{\vs\sim \rho^{\pi_b}(\vs)} \sbr{\KL(\pi_b(\cdot\vert\vs)\|\mu(\cdot\vert\vs))} \leq \rvepsilon,\\
                  & \int_\va \mu(\va\vert\vs)d\va =1,\quad \forall \vs,
         \label{den}
\end{align}

where $\mu(\va\vert\vs)$ denotes our diffusion-based policy, $\pi_b$ denotes the behavior policy. Here we represent our policy $\mu(\va\vert\vs)$ via the conditional diffusion model:
\begin{equation}
\begin{aligned}
        \mu(\va\vert\vs) &= \int \mu(\va^{0:T}\vert\vs)d\va^{1:T}\\
        &=\int\gN (\va^T; \mathbf{0},\mI)\prod_{i=1}^T\mu(\va^{i-1}\vert\va^{i},\vs)d\va^{1:T},
\end{aligned}
    \label{policy}
\end{equation}
where the end sample of the reverse chain $\va^0$ is the action used for the policy's output and $\mu(\va^{0:T})$ is the joint distribution of all noisy samples. According to DDPM, we can approximate the reverse process $\mu(\va^{i-1}\vert\va^{i},\vs)$ with a Gaussian distribution $\gN(\va^{i-1}; \vmu_{\theta}(\va^{i}, \vs, i), \mSigma_{\theta}(\va^{i}, \vs, i))$.
We also follow the DDPM to fix the covariance matrix and predict the mean with a conditional noise model $\rvepsilon_\theta(\va^{i}, \vs, i)$:
\begin{equation}
        \textstyle\vmu_{\theta}(\va^{i}, \vs, i) = \frac{1}{\sqrt{\alpha_i}} \big( \va^{i} - \frac{\beta_i}{\sqrt{1 - \bar{\alpha}_i}} \rvepsilon_\theta(\va^{i}, \vs, i) \big).
\end{equation}
In the reverse process,  we sample $\va^T \sim \gN(\mathbf{0}, \mI)$ and then %
    follow the reverse diffusion chain $\gN\cbr{\va^{i-1} \vert \va^i}$, parameterized by $\theta$ as %
 \begin{equation} \label{reverse_sampling}
        \gN\cbr{a^{i-1};\frac{\va^i}{\sqrt{\alpha_i}} - \frac{\beta_i}{\sqrt{\alpha_i(1 - \bar{\alpha}_i)}} \rvepsilon_\theta(\va^{i}, \vs, I),\beta_i }
\end{equation}
for $i=T,\ldots,1.$  The reverse sampling in \Eqref{reverse_sampling} requires iteratively predicting $\rvepsilon$ $T$ times. When $T$ is large, it will consume much time during the sampling process. To work with small $T$ ($T=5$ in our experiment), we follow \citep{xiao2021,wang2022b} and define
   \begin{equation*}
        \beta_i = 1 - \alpha_i = 1 - e^{-\beta_{\min}(\frac{1}{T}) - 0.5(\beta_{\max} - \beta_{\min})\frac{2i -1}{T^2}},
    \end{equation*}
which is a noise schedule obtained under the variance preserving SDE of~\citet{song2020}.

\begin{theorem}
\label{theorem1}
Let $\mu(\va\vert\vs)$ be a diffusion-based policy and $\pi_b$ be the behavior policy. Then, we have
\begin{enumerate}
\item[(1)] There exists  $\rvkappa_0\in\gR$ such that \Eqref{kl} can be transformed to
$$H(\pi_b,\mu)=-\E_{\vs\sim \rho^{\pi_b}(\vs),\va\sim\pi_b(\cdot\vert\vs)}\sbr{\log\mu(\cdot\vert\vs)} \leq \rvkappa_0.$$
    \item[(2)] $\forall s$, $\int_\va\mu(\va\vert\vs)d\va\equiv1$.
\end{enumerate}
\end{theorem}

\begin{corollary}
\label{corollary1}
    The primal problem~(\Eqref{obj}-\Eqref{den}) can be transformed into the following DiffCPS problem:
    \begin{align}
        \mu^{*} = \argmax_\mu \ &J(\mu) = \argmax_\mu \int_\mathcal{S} d_0(\vs) \int_\mathcal{A}Q^\mu(\vs, \va) d\va d\vs \nonumber \\
         s.t. \quad &     H(\pi_b,\mu) \leq \rvkappa_0.     \label{cps}
    \end{align}
\end{corollary}
\begin{remark}
    Theorem~\ref{theorem1} and Corollary~\ref{corollary1} formulate our diffusion-based CPS problem, then we will show under some mild assumption, the strong duality holds for \Eqref{cps}.
\end{remark}
\begin{assumption}
    \label{ass1}
    Suppose that $r$ is bounded and that Slaters' condition \citep{boyd2004convex} holds for our offline RL setting. 
\end{assumption}
\begin{assumption}
    \label{ass2}
    Suppose that all policies learned from the offline dataset $\gD$ satisfy 
    \begin{equation}
    \label{eq15}
        \mu(\va\vert\vs)=\frac{\rho^\mu(\vs,\va)}{\rho^{\pi_b(\vs)}(\vs)},
    \end{equation} 
    where $\rho^{\pi_b}(\vs)$ denotes the normalized discounted state visitation frequencies induced by the behavior policy.
\end{assumption}
\begin{remark}
    Assumption \ref{ass2} is a mild assumption in offline reinforcement learning, as the state distributions encountered by all policies can only be those collected by the behavior policy. Therefore, it is feasible to assume \Eqref{eq15} regarding the occupancy measure of the learned policy.
\end{remark}

\begin{theorem}
\label{theorem2}
Under Assumption \ref{ass1} and Assumption \ref{ass2}, strong duality holds for \Eqref{cps}, which has the same optimal policy $\mu^*$ with \Eqref{eq:minmax}.
\end{theorem}
    According to Theorem \ref{theorem2}, we can get the optimal policy by solving the dual problem. In the following section, we outline the algorithm for solving \Eqref{cps}. It consists of two steps: (i) solving the unconstrained dual function of \Eqref{cps}, (ii) updating the Lagrange multiplier. 

\subsection{Primal-Dual Method for DiffCPS}\label{pd-solution-diffcps}

In this section, the duality is used to solve the \Eqref{cps} and derive our method. The dual function of \Eqref{cps} is
\begin{equation}
\begin{aligned}
    & \quad d(\lambda)=\argmax_{\mu}\gL(\mu,\lambda)\\
    &\text{where} \quad \gL(\mu,\lambda)=J(\mu)+\lambda(\rvkappa_0-H(\pi_b,\mu)).
\end{aligned}
    \label{mu}
\end{equation}
The Lagrange dual problem associated with the \Eqref{cps} is
\begin{equation}
\label{eq:minmax}
    D^{*}(\lambda)\triangleq\min_{\lambda\geq 0 }\max_\mu J(\mu)+\lambda(\rvkappa_0-H(\pi_b,\mu)).
\end{equation}
In \Eqref{mu}, we need to calculate the cross entropy $H(\pi_b,\mu)$, which is intractable in practice.

\begin{proposition}
\label{proposition1}
    In the diffusion model, we can approximate the entropy with MSE-like loss $\gL_c(\pi_b,\mu)$ through ELBO:
\begin{equation}
H(\pi_b,\mu)\approx c+\gL_c(\pi_b,\mu),
\label{constraint2}
\end{equation}
where c is a constant.
\end{proposition}
\begin{remark}
    Let $\rvkappa=\rvkappa_0-c$, Proposition~\ref{proposition1} allow us to use the loss of diffusion model to approximate the \Eqref{kl}. In the following section, we will replace the  $H(\pi_b,\mu)$ with $\gL_c(\pi_b,\mu)$.
\end{remark}
In practice, to turn the original functional optimization problem into a tractable problem, we often use parametrized policies, such as neural networks. Next, we shift focus to solving the parametric version of \Eqref{cps}, $i.e.$ to find the parametrized policies $\mu_\theta$ that solve \Eqref{mu}
\begin{equation}
\begin{aligned}
    &\quad d_\theta(\lambda)\triangleq\max_{\theta} \gL(\mu_\theta,\lambda_t)\\
    & \text{where} \quad \gL(\mu_\theta,\lambda_t)=\E_{\vs\sim d_0(\vs),\va\sim\mu_\theta(\cdot\vert\vs)}\sbr{\frac{Q^{\mu_\theta}(\vs,\va)}{Q_\text{target}^{\mu_\theta}(\vs,\va)}}\\&+\lambda(\rvkappa-\gL_c(\pi_b,\mu_\theta^{*})),
\end{aligned}
    \label{mu2}
\end{equation}
\begin{equation}
D_\theta^{*}\triangleq \min_{\lambda\geq0}d_\theta(\lambda).
    \label{eta2}
\end{equation}
Notice that we follow~\citet{fujimoto2021,wang2022b} to normalize $Q^{\mu}(\vs,\va)$ by dividing it by the target Q-net's value. We also clip the $\lambda$ to keep the constraint $\lambda \geq 0$ holding by $\lambda_\text{clip}=\max(c,\lambda)$, $c\geq0$. We also find that we can improve the performance of the policy by delaying the policy update~\citep{fujimoto2018}.  The algorithm given by \Eqref{mu2}-\Eqref{eta2} for solving \Eqref{cps} is summarized under \Algref{alg:main}. 

\begin{algorithm}[t]
\small
    \caption{DiffCPS}
    \begin{algorithmic}[1]
    \label{alg:main}
    \STATE Initialize policy network $\mu_{\theta}$, critic networks $Q_{\phi_1}$, $Q_{\phi_2}$, and target networks $\mu_{\theta'}$, $Q_{\phi_1'}$, $Q_{\phi_2'}$,\!
    \STATE policy evaluation interval  $d$ and step size $\eta$.
    \FOR{$t=1$ to $T$}
    \STATE Sample transition mini-batch $\gB\!=\!\cbr{(\vs_t, \va_t, r_t, \vs_{t+1})}\!\sim\!\gD$.
    \STATE \textit{\bfseries \# Critic updating}
    \STATE Sample $\va_{t+1}\sim\mu_\theta(\cdot\vert\vs_{t+1})$ according to \Eqref{reverse_sampling}.
    \STATE $y=r_t+\gamma \min_{i=1,2}Q_{\phi_i'}\cbr{\vs_{t+1},\va_{t+1}}$.
    \STATE Update critic $\phi_i \leftarrow \argmin_{\phi_i} N^{-1}\sum(y-Q_{\phi_i}(\vs_t,\va_t))^2$.
    \IF{$t$ mod $d$} 
    \STATE Sample $\va\sim\mu_\theta(\cdot\vert\vs)$ according to \Eqref{reverse_sampling}. 
    \STATE \textit{\bfseries \# Policy updating through \Eqref{mu2}}
    \STATE $\theta_{t+1}\approx\argmax \gL(\mu_\theta,\lambda_t)$.
    \STATE \textit{\bfseries \# Lagrange multiplier $\lambda$ updating through \Eqref{eta2}}
    \STATE $\lambda_{t+1}\leftarrow\lambda_t-\eta\sbr{\rvkappa-\gL_c(\pi_b,\mu^{*}_{\theta_{t+1}})}$.
    \STATE $\lambda_{t+1}\leftarrow \lambda_\text{clip}=\max(c,\lambda_{t+1})$, $c\geq0$.
    \ENDIF
    \STATE \textit{\bfseries \# Target Networks updating}
        \STATE $\theta' = \rho \theta' + (1 - \rho) \theta_{t+1}$, $\phi_i' = \rho \phi_i' + (1 - \rho) \phi_i \mbox{ for } i=\{1,2\}$.
    \ENDFOR
    \end{algorithmic}
    \end{algorithm}
    
However, there is a cost for introducing such parametrization in \Algref{alg:main}: the duality gap is no longer null. In the following Theorem~\ref{T:convergence}, we show that after introducing parameterization, we can obtain a suboptimal solution in a $\gO(1/\epsilon)$ number of dual iterations. Below, we first define $\epsilon$-universal parametrization functions.

\begin{definition}\label{def_universal_param}
  A parametrization $\pi_\theta$ is an $\epsilon$-universal parametrization of functions in $\gP(\gS)$ if, for some $\epsilon >0$, there exists for any $\pi\in\gP(\gS)$ a parameter $\theta \in \mathbb{R}^p$ such that
  \begin{equation}
\max_{s\in\gS}\int_{\gA}\left|\pi(a|s)-\pi_\theta(a|s)\right|\,da\leq \epsilon.
  \end{equation}  
  \end{definition}

\begin{table*}[t]
\caption{The performance of DiffCPS and other SOTA baselines on D4RL tasks. The mean and standard deviation of DiffCPS are obtained by evaluating the trained policy on five different random seeds. We report the performance of baseline methods using the best results reported from their paper. ``-A" refers to any number of hyperparameters allowed. Results within $3$ percent of the maximum in every D4RL task and the best average result are highlighted in boldface.}
\label{tbl:rl_results}
\vspace{0.2cm}
\centering
\small
\resizebox{1.0\textwidth}{!}{%
\begin{tabular}{llccccccccc}
\toprule
\multicolumn{1}{c}{\bf Dataset} & \multicolumn{1}{c}{\bf Environment} & \multicolumn{1}{c}{\bf CQL}& \multicolumn{1}{c}{\bf IDQL-A}  & \multicolumn{1}{c}{\bf QGPO} & \multicolumn{1}{c}{\bf SfBC} & \multicolumn{1}{c}{\bf DD} & \multicolumn{1}{c}{\bf Diffuser} & \multicolumn{1}{c}{\bf Diffuison-QL}  & \multicolumn{1}{c}{\bf IQL}  & \multicolumn{1}{c}{\bf DiffCPS(ours)}\\
\midrule
Medium-Expert & HalfCheetah    & $62.4$               &  $\bf{95.9}$ &  $93.5   $  & $92.6$   & $90.6$     & $79.8$       &  $\bf{96.8}$& $ 86.7$ & $\bf{100.3\pm4.1}$               \\
Medium-Expert & Hopper         & $98.7$               &  $108.6$    &  $    108.0$ & $108.6$  &$\bf{111.8}$& $107.2$ &$\bf{111.1}$ & $91.5$      & $\bf{112.1\pm0.6}$            \\
Medium-Expert & Walker2d       & $110.1$         & $\bf{112.7}$      & $\bf{110.7}$& $109.8$  &$108.8$& $108.4$ & $110.1$& $109.6$ & $\bf{113.1\pm1.8}$            \\
\midrule
Medium        & HalfCheetah    &  $44.4$              &  $51.0$     &  $\bf{54.1}     $& $45.9$        & $49.1$     & $44.2$       &  $\bf{51.1}$     & $47.4$       & $\bf{71.0\pm0.5}$            \\
Medium        & Hopper         &  $58.0$              &  $65.4$     &  $\bf{98.0}$     & $57.1$        & $79.3$     & $58.5$       &$ \bf{90.5}      $& $66.3$       & $\bf{100.1\pm3.5}$  \\
Medium        & Walker2d        &  $79.2$              &  $82.5$     &  $\bf{86.0}$     & $77.9$        & $82.5$ & $79.7$       &$\bf{87.0}$  & $78.3$       & $\bf{90.9\pm1.6}$  \\
\midrule
Medium-Replay & HalfCheetah    &  $46.2$         &  $45.9$     &  $\bf{47.6}$     &   $37.1$      &$39.3$& $42.2$       &$\bf{47.8}$  & $44.2$       & $\bf{50.5\pm0.6}$  \\
Medium-Replay & Hopper         &  $48.6$              &  $92.1$     &  $96.9$     &   $86.2$      & $\bf{100}$     & $96.8$  & $\bf{101.3}$& $94.7$       & $\bf{101.1\pm0.2}$  \\
Medium-Replay & Walker2d       &  $26.7$              &  $\bf{85.1}$     &  $84.4$     &   $65.1$      & $75.0$     & $61.2$       &  $\bf{95.5}$& $73.9$       & $\bf{91.3\pm0.7}$  \\
\midrule
\multicolumn{2}{c}{\bf Average (Locomotion)}&$63.9$   &  $82.1$     & $86.6$      &   $75.6$      &$81.8$      &      $75.3$       & $87.9$ & $76.9$       & $\bf{92.26}$  \\
\midrule
Default       & AntMaze-umaze  &  $74.0$              &  $\bf{94.0}$     & $\bf{96.4}$      & $92.0$   & -          & -            & $93.4$      & $87.5$       & $\bf{97.4\pm3.7}$  \\
Diverse       & AntMaze-umaze  &  $\bf{84.0}$         &  $80.2  $   & $74.4$      & $\bf{85.3}$   & -          & -            &$66.2$       & $62.2$       & $\bf{87.4\pm3.8}$  \\
\midrule  
Play          & AntMaze-medium &  $61.2$              &  $\bf{84.5}$      & $\bf{83.6}$      & $81.3$   & -          & -            & $76.6$       & $71.2$        & $\bf{88.2\pm2.2}$  \\
Diverse       & AntMaze-medium &  $53.7$              &  $\bf{84.8}$      & $\bf{83.8}$      & $82.0$   & -          & -            & $78.6$      & $70.0$       & $\bf{87.8\pm6.5}$  \\
\midrule
Play          & AntMaze-large  &  $15.8$              &  $\bf{63.5}$      & $\bf{66.6}$      & $59.3$        & -          & -            & $46.4$      & $39.6$       & $\bf{65.6\pm3.6}$  \\
Diverse       & AntMaze-large  &  $14.9$              &  $\bf{67.9}$      & $\bf{64.8}$      & $45.5$        & -          & -            & $57.3$       &  $47.5$       & $\bf{63.6\pm3.9}$  \\
\midrule
\multicolumn{2}{c}{\bf Average (AntMaze)}&  $50.6$    &  $79.1$     & $78.3$      &   $74.2$      & -            &   -   &  $69.8$    & $63.0$       & $\bf{81.67}$  \\
\midrule
\multicolumn{2}{c}{\bf{\# Diffusion steps}}&-   &  $5$     & $15$  & $15$  &   $100$     & $100$  &   $5$   & -       & $5$ \\
\bottomrule
\end{tabular}
}
\end{table*}

\begin{theorem}\label{T:convergence}
  Let $\pi_\theta$ be an $\epsilon$ universal parametrization of $\gP(\gS)$ according to Definition \ref{def_universal_param}, Suppose that~$r$ is bounded  by constants~$R>0$ and $B = \max_{\vs,\va}\rho^{\pi_b}(\vs,\va)1/\rho(\vs,\va)$. and~$\gamma\in(0,1)$ be the discount factor. Then, under Assumption \ref{ass1} \ref{ass2} and for any $\varepsilon>0$, the sequence of updates of Algorithm \ref{alg:main} with step size $\eta$ converges in $K>0$ steps, with 
 \begin{equation}
          K \leq {\left\|\lambda_0-\lambda_\theta^\star\right\|^2}{\big/}{2\eta \varepsilon},
 \end{equation} 
 to a neighborhood of $P^\star$ (the solution of \Eqref{cps}), satisfying
\begin{equation}\label{eqn_convergence_neighborhood}
\!\!\!\!-\frac{\left(R+\left\|\lambda^\star\right\|_1B\right)\epsilon}{1-\gamma}
\!\leq\!  d_{\theta}(\lambda_K) \!-\!  P^\star
\!\leq\! \eta\frac{C}{2}+\delta+\varepsilon.
  \end{equation}
where~$C =\left( \rvkappa_0+\int_{\gS\times\gA}\rho^{\pi_b}(\vs,\va)\log\pi_b(\va\vert\vs)d\va d\vs\right)^2$, $\delta$ denotes the error between the parameterized local optimal solution and the optimal solution in \Eqref{mu2}.  and~$\lambda^\star$ be the solution to the dual problem associated with \Eqref{cps}.
\end{theorem}
\begin{remark}
    Notice that the suboptimal solution which the primal-dual algorithm convergences depends on the goodness of the solution of \Eqref{mu2} and the representation ability of the parametrized policy, while the expressiveness of the model is precisely the greatest strength of diffusion models.
\end{remark}
Theoretically, we need to precisely solve the \Eqref{mu2} and \Eqref{eta2}. However, in practice, we can resort to SGD and parametrized policy to solve the equations. In this way, we can recursively optimize \Eqref{cps} through \Eqref{mu2} and \Eqref{eta2}. The policy improvement described in \Eqref{mu2} and the solution of \Eqref{eta2} constitute the core of DiffCPS.

\section{Experiments}
We evaluate our DiffCPS on D4RL~\citep{fu2020} benchmark in Section~\ref{d4rl}. Further, we conduct an ablation experiment to assess the contribution of different parts in DiffCPS in Section~\ref{ablation}. Due to space limitations, we place the experiment setup, implementation details, and additional experiment results in the \textbf{Appendix}~\ref{details}.

\subsection{Results}
\label{d4rl}

In Table \ref{tbl:rl_results}, we compare the performance of DiffCPS to other offline RL methods in D4RL~\citep{fu2020} tasks. We merely illustrate the results of the MuJoCo locomotion and AntMaze tasks due to the page limit. In traditional MuJoCo tasks, DiffCPS outperforms other methods as well as recent diffusion-based method~\citep{wang2022b,lu2023,hansen-estruch2023} by large margins in most tasks, especially in the HalfCheetah medium. The medium datasets are collected by online SAC~\citep{haarnoja2018} agent trained to approximate $1/3$ the performance of the expert. Hence, it contains a lot of suboptimal trajectories, which makes the offline RL algorithms hard to learn.

Compared to the locomotion tasks, AntMaze tasks are more challenging since the datasets consist of sparse rewards and suboptimal trajectories. Even so, DiffCPS also achieves competitive or SOTA results compared with other methods. For simpler tasks like umaze, DiffCPS achieves  $100\pm 0\% $ success rate on some seeds, which shows its powerful ability to learn from suboptimal trajectories. In other AntMaze tasks, DiffCPS also shows competitive performance compared to other SOTA diffusion-based approaches. Overall, the experiments demonstrate the effectiveness of DiffCPS compared to several existing methods for offline RL.

\begin{table*}[!t]
\caption{Ablation study of diffusion steps. We conduct an ablation study to investigate the impact of diffusion steps on different algorithms. We only show the average score due to the page limit. The results of DiffCPS are obtained from three random seeds, while the results of SfBC are derived from the original SfBC paper.}
\label{table:ablation}
\vspace{0.1in}
\centering
\small
\scalebox{0.8}{
\begin{tabular}{lccccccc}
\toprule
\multicolumn{1}{c}{\bf D4RL Tasks}  & \multicolumn{1}{c}{\bf DiffCPS (T=5) }
& \multicolumn{1}{c}{\bf DiffCPS (T=15) }& \multicolumn{1}{c}{\bf DiffCPS (T=20) }
& \multicolumn{1}{c}{\bf SfBC (T=10)} &\multicolumn{1}{c}{\bf SfBC (T=15)} &\multicolumn{1}{c}{\bf SfBC (T=25)}  \\
\midrule
\multicolumn{1}{c}{\bf Locomotion}  &      $\bf{92.0}$       &  $87.5$   &  $87.6$ &  $72.9$ &  $\bf{75.6}$ &  $74.4$    \\

\midrule
\multicolumn{1}{c}{\bf AntMaze} &      $\bf{80.0}$       &  $60.7$ &    $66.7$ &  $65.7$ &  $\bf{74.2}$ &  $73.0$    \\

\bottomrule
\end{tabular}
}
\end{table*}

\begin{figure*}[!t]
\centering
\includegraphics[width = .9\linewidth]{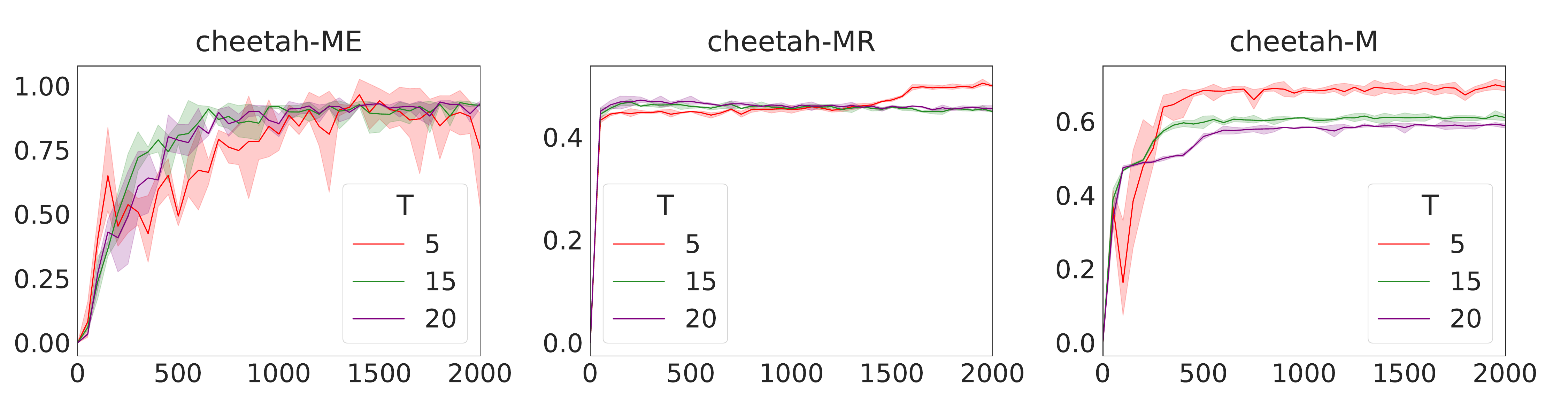}
\vspace{-0.2in}
\caption{Ablation studies of diffusion steps $T$ on selected Gym tasks (three random seeds). We observe that as $T$ increases, the training stability improves, but the final performance drops.}
\label{fig:ablationT}
\label{ablation_T}
\end{figure*}

\subsection{Ablation Study}
\label{ablation}
In this section, we analyze why DiffCPS outperforms the other methods quantitatively on D4RL tasks. We conduct an ablation study to investigate the impact of three parts in DiffCPS, $ i.e.$ diffusion steps, the minimum value of Lagrange multiplier $\lambda_{\text{clip}}$, and policy evaluation interval.

\textbf{Diffusion Steps.} We show the effect of diffusion steps $T$, which is a vital hyperparameter in all diffusion-based methods. In SfBC the best $T$ is $15$, while $T=5$ is best for our DiffCPS. Table \ref{table:ablation} shows the average performance of different diffusion steps. Figure \ref{ablation_T} shows the training curve of selected D4RL tasks over different diffusion steps $T$.

We also note that large $T$ works better for the bandit experiment. However, for D4RL tasks, a large $T$ will lead to a performance drop. The reason is that compared to bandit tasks, D4RL datasets contain a significant amount of suboptimal trajectories. A larger $T$ implies stronger behavior cloning ability, which can indeed lead to policy overfitting to the suboptimal data, especially when combined with actor-critic methods. Poor policies result in error value estimates, and vice versa, creating a vicious cycle that leads to a drop in policy performance with a large $T$. 

\begin{figure*}[!t]
\centering
\includegraphics[width = .8\linewidth]{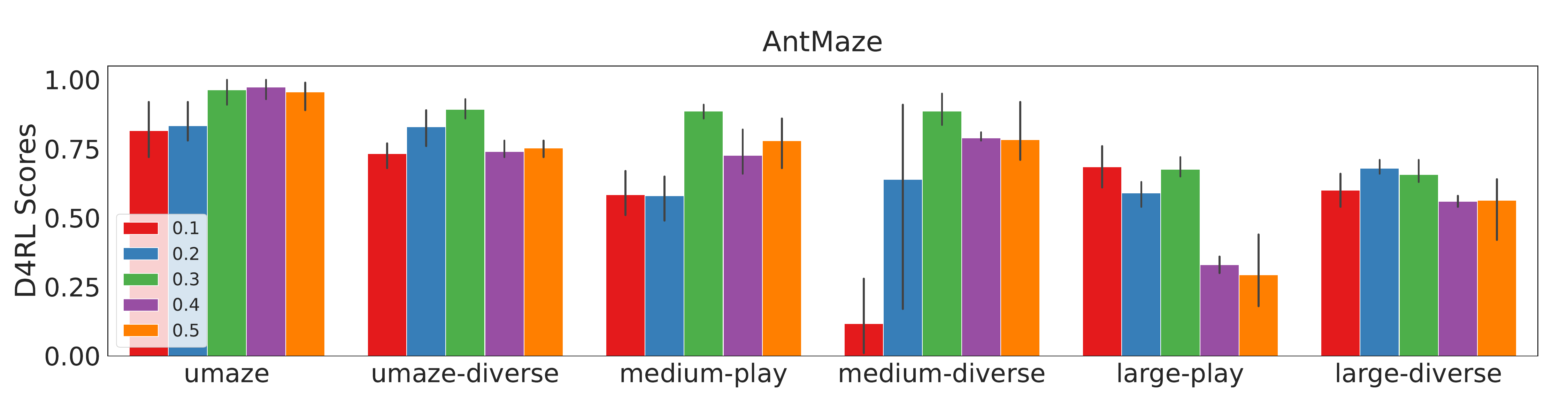}\\
\includegraphics[width = .8\linewidth]{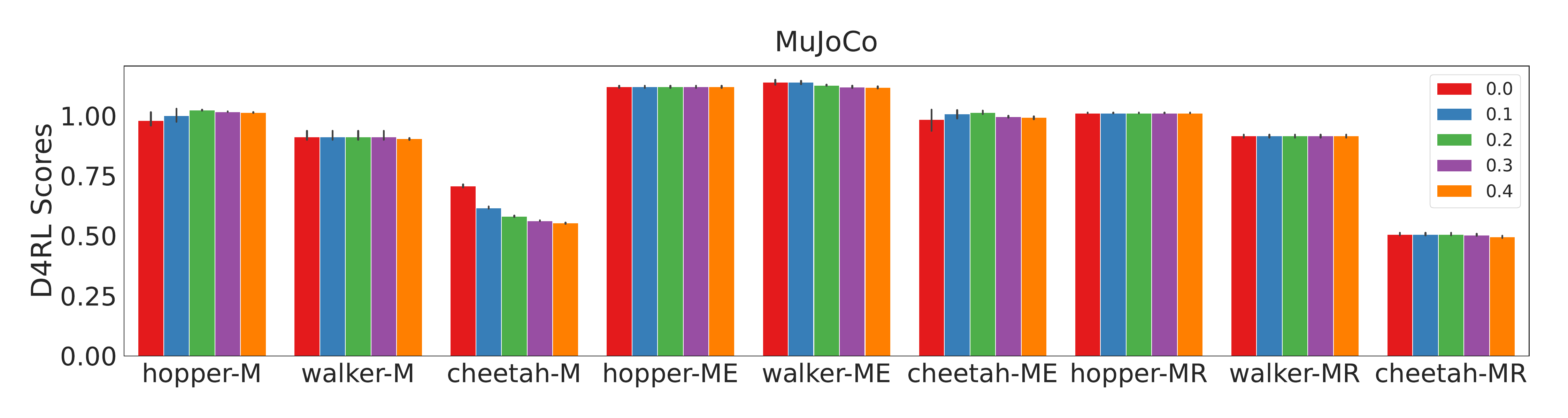}
\vspace{-0.2in}
\caption{Ablation studies of $\lambda_{\text{clip}}$ in AntMaze and MuJoCo tasks. We observe that $\lambda_{\text{clip}}$ has little impact on MuJoCo tasks but significantly influences AntMaze tasks, especially as AntMaze datasets are larger. The reason is that the sparse rewards and suboptimal trajectories in AntMaze datasets make the critic network prone to error estimation, leading to learning poor policy. Therefore, there is a need to enhance learning from the original dataset which means we should increase $\lambda$ or enhance the KL constraint. We find that increasing $\lambda_{\text{clip}}$ while maintaining a moderate KL constraint achieves the best results. All the results are obtained by evaluating three random seeds. }
\label{fig:ablationeta}
\end{figure*}

\textbf{The Minimum Value of Lagrange multiplier $\lambda_{\text{clip}}$.} In our method, $\lambda$ serves as the coefficient for the policy constraint term, where a larger $\lambda$ implies a stronger policy constraint. Although we need to restrict the $\lambda\geq0$ according to the definition of Lagrange multiplier, we notice that we could get better results through clip $\lambda\geq c$ in AntMaze tasks, where $c$ is a positive number, see full $\lambda$ ablation results in Figure \ref{fig:ablationeta} for details.

\begin{figure*}[t]
\centering
\includegraphics[width = .9\linewidth]{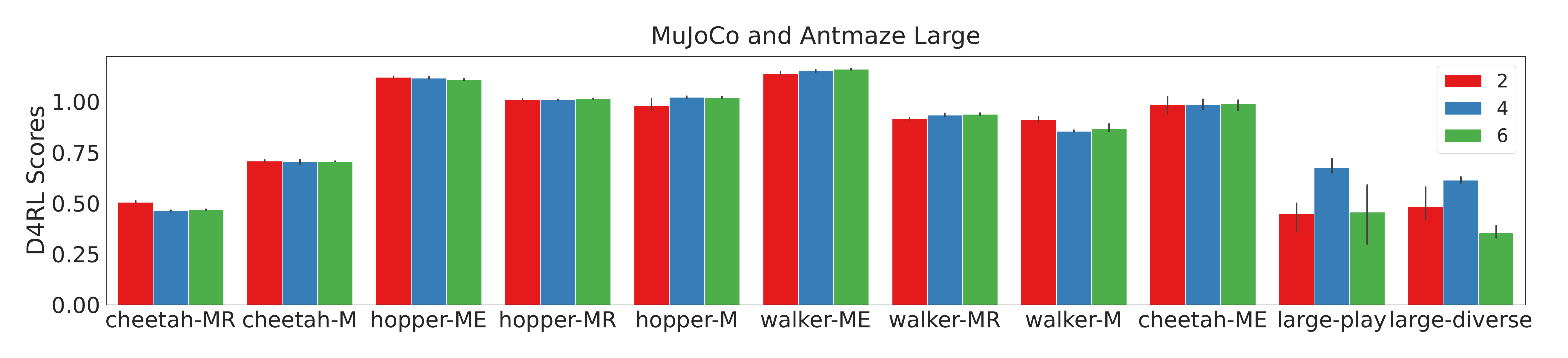}\\
\vspace{-0.2in}
\caption{Ablation studies of the policy evaluation interval in AntMaze and MuJoCo tasks. Delayed policy updates have a relatively minor impact on the MuJoCo locomotion tasks. However, for large-scale sparse reward datasets like AntMaze Large, choosing an appropriate update frequency can greatly increase the final optimal results. The MuJoCo task results are obtained with 2 million training steps (three random seeds), while AntMaze results are obtained with 1 million training steps (three random seeds). }
\label{fig:ablationfreq}
\end{figure*}
\textbf{Policy Evaluation Interval.} We include the ablation of policy evaluation interval in Figure \ref{fig:ablationfreq}. We find that the policy delayed update~\citep{fujimoto2018} has significant impacts on AntMaze large tasks. However, for other tasks, it does not have much effect and even leads to a slight performance decline. The reason is that infrequent policy updates reduce the variance of value estimates, which is more effective in tasks where sparse rewards and suboptimal trajectories lead to significant errors in value function estimation like AntMaze large tasks.

In a nutshell, the ablation study shows that the combination of these three parts, along with DiffCPS, collectively leads to producing good performance.
\vspace{-2mm}

\section{Related Work}
\label{gen_inst}

\textbf{Offline Reinforcement Learning}. Offline RL algorithms need to avoid extrapolation error. Prior works usually solve this problem through policy regularization~\citep{fujimoto2019,kumar2019stabilizing,wu2019behavior,fujimoto2021}, value pessimism about unseen actions~\citep{kumar2020,kostrikov2021offline}, or implicit TD backups~\citep{kostrikov2021,ma2021offline} to avoid the use of out-of-distribution actions. Another line of research solves the offline RL problem through weighted regression~\citep{peters2010,peng2019,nair2020} from the perspective of CPS. Our DiffCPS derivation is related but features with a diffusion model form.

\textbf{Diffusion Models in RL}. There exist several works that introduce the diffusion model to RL. Diffuser~\citep{janner2022} uses the diffusion model to directly generate trajectory guided with gradient guidance or reward. DiffusionQL~\citep{wang2022b} uses the diffusion model as an actor and optimizes it through the TD3+BC-style objective with a coefficient $\eta$ to balance the two terms. AdaptDiffuser~\citep{liang2023} uses a diffusion model to generate extra trajectories and a discriminator to select desired data to add to the training set to enhance the adaptability of the diffusion model. DD~\citep{ajay2022}  uses a conditional diffusion model to generate trajectory and compose skills. Unlike Diffuser, DD diffuses only states and trains inverse dynamics to predict actions. QGPO~\citep{lu2023} uses the energy function to guide the sampling process and proves that the proposed CEP training method can get an unbiased estimation of the gradient of the energy function under unlimited model capacity and data samples.  IDQL~\citep{hansen-estruch2023} reinterpret IQL as an Actor-Critic method and extract the policy through sampling from a diffusion-parameterized behavior policy with weights computed from the IQL-style critic. EDP~\citep{kang_efficient_nodate} focuses on boosting sampling speed through approximated actions. SRPO~\citep{chen_score_2023} uses a Gaussian policy in which the gradient is regularized by a pretrained diffusion model to recover the IQL-style policy. DiffCPS is distinct from these methods because we derive it from \eqref{cps}.

The closest to our work is the method that combines AWR and diffusion models. SfBC~\citep{chen2022} uses the diffusion model to generate candidate actions and uses the regression weights to select the best action. Our method differs from it as we directly solve the limited policy expressivity problem through the primal-dual method for DiffCPS without resorting to AWR. This makes DiffCPS simple to implement and tune hyperparameters. As shown in Table \ref{tab:hyperparameter} we can achieve SOTA results in most tasks by merely tuning one hyperparameter.
 
\section{Conclusion}
In our work, we solve the limited expressivity problem in the weighted regression through the diffusion model. We first simplify the CPS problem with the action distribution of diffusion-based policy. Then we prove that strong duality holds for diffusion-based CPS problems, and solve it through the primal-dual method with function approximation. Experimental results on the D4RL benchmark illustrate the superiority of our method which outperforms previous SOTA algorithms in most tasks, and DiffCPS is easy to tune hyperparameters, which only needs to tune the constraint $\rvkappa$ in most tasks. We hope that our work can inspire relative researchers to utilize powerful generative models, especially the diffusion model, for offline RL and decision-making.
\section{Impact statements}
Offline RL aims to learn a policy from an offline dataset, much like supervised learning, but due to extrapolation error and function approximation, the task of offline reinforcement learning is much more challenging compared to supervised learning and pre-training. The DiffCPS algorithm proposed in this paper utilizes diffusion-based policy and primal-dual method to address the CPS problem in offline reinforcement learning. Our method will promote the development of the field of offline reinforcement learning, thus facilitating the implementation of offline reinforcement learning in practical scenarios, such as robotic control, and consequently bring about a series of corresponding ethical issues and social outcomes.

\bibliography{example_paper}
\bibliographystyle{icml2024}

\newpage
\appendix
\onecolumn

In the appendix, we provide the proofs of the main theorems in the paper~(Appendix~\ref{proof}), the notation table~(Appendix~\ref{notation}), details of the experiments~(Appendix~\ref{details}), some additional experiments~(Appendix~\ref{toy_append},\ref{extra experiments},\ref{extra_related}), and a discussion on the use of the augmented Lagrangian method~(Appendix~\ref{alm}).

\section{Proofs}
\label{proof}
\subsection{Theorem~\ref{theorem1}}
We provide the proof of Theorem~\ref{theorem1}.
\begin{proof}
(1): Since the pointwise KL constraint in \Eqref{origin_cps}  at all states is intractable, we follow the \citet{peng2019} to relax the constraint by enforcing it only in expectation $$E_{\vs\sim \rho^{\pi_b}(\vs)} \sbr{\KL(\pi_b(\cdot\vert\vs)\|\mu(\cdot\vert\vs))}\leq \rvepsilon.$$
The LHS can be transformed to
\begin{equation}
    E_{\vs\sim \rho^{\pi_b}(\vs)} \sbr{\KL(\pi_b(\cdot\vert\vs)\|\mu(\cdot\vert\vs))} =E_{\vs\sim \rho^{\pi_b}(\vs),\va\sim\pi_b(\cdot\vert\vs)}\sbr{\log\pi_b(\cdot\vert\vs)-\log\mu(\cdot\vert\vs)},
\end{equation}
where 
$$E_{\vs\sim \rho^{\pi_b}(\vs),\va\sim\pi_b(\cdot\vert\vs)}\sbr{\log\pi_b(\cdot\vert\vs)}$$ is the negative entropy of  $\pi_b$ is a constant for $\mu$.
So the constraint in \Eqref{kl} can be transformed to
\begin{equation}
H(\pi_b,\mu)=-\E_{\vs\sim \rho^{\pi_b}(\vs),\va\sim\pi_b(\cdot\vert\vs)}\sbr{\log\mu(\cdot\vert\vs)} \leq \rvkappa_0,
    \label{constraint1}
\end{equation}
where $\rvkappa_0=\rvepsilon-E_{\vs\sim \rho^{\pi_b}(\vs),\va\sim\pi_b(\cdot\vert\vs)}\sbr{\log\pi_b(\cdot\vert\vs)}$. 

(2): According to the definition of the diffusion model in \Eqref{policy}:
\begin{equation}
    \begin{aligned}
        \int_\va\mu(\va\vert\vs)d\va
        &=\int_\va\mu(\va^0\vert\vs)d\va^0\\
        &= \int\int \mu(\va^{0:T}\vert\vs)d\va^{1:T}d\va^0 \\
        &=\int \mu(\va^{0:T})d\va^{0:T}\\
        &= 1.
    \end{aligned}
    \label{proof4}
\end{equation}
\Eqref{proof4} implies the constraint \Eqref{den} always holds for the diffusion model, so we can omit this constraint.

According to the above two properties, we can rewrite the diffusion-based CPS problem \Eqref{obj}-\Eqref{den} as 
\begin{equation}
    \begin{aligned}
        \mu^{*} = \argmax_\mu \ &J(\mu) = \argmax_\mu \int_\mathcal{S} d_0(\vs) \int_\mathcal{A}Q^\mu(\vs, \va) d\va d\vs\\
         s.t. \quad &     H(\pi_b,\mu) \leq \rvkappa_0.
    \end{aligned}
    \label{cps_proof}
\end{equation}
\end{proof}

\subsection{Theorem~\ref{theorem2}}
Then we prove the Theorem~\ref{theorem2}.

The perturbation function associated to \Eqref{cps_proof} is defined as
\begin{equation}
    \begin{aligned}
        P(\xi)= &\max_\mu \gJ(\mu)=\max_\mu \int_\mathcal{S} d_0(\vs) \int_\mathcal{A}Q^\mu(\vs, \va) d\va d\vs\\
         s.t. \quad &     -H(\pi_b,\mu)\geq -\rvkappa_0+\xi.
    \end{aligned}
    \label{perturbation}
\end{equation}

\begin{lemma}\label{T:strongDuality}
If (i)~$r$ is bounded; (ii)~Slater's condition holds for~\Eqref{cps_proof} and (iii)~its perturbation function~$P(\xi)$ is concave, then strong duality holds for~\Eqref{cps_proof}.
\end{lemma}
\begin{proof}
    See, e.g., \citet{rockafellar1970convex}[Cor.~30.2.2]
\end{proof}
\begin{proof}

 (The proof skeleton is essentially based on \citet{paternain2019constrained} Theorem 1)
 
Condition (i) and (ii) are satisfied by the Assumption \ref{ass1}. To prove the strong duality of \Eqref{cps_proof}, it suffices then to show the perturbation function is concave [(iii)], i.e., that for every $\xi^1,\xi^2\in\R$, and $t\in(0,1)$,
\begin{equation}
    \label{concave}
	P\left[ t \xi^1 + (1-t) \xi^2 \right]
		\geq t P\left( \xi^1 \right) + (1-t) P\left( \xi^2 \right)
		\text{.}
\end{equation}

    If for either perturbation $\xi^1$ or $\xi^2$ the problem becomes infeasible then $P(\xi^1)=-\infty$ or $P(\xi^2)=-\infty$ and thus \Eqref{concave} holds trivially. For perturbations which keep the problem feasible, suppose~$P(\xi^1)$ and~$P(\xi^2)$ are achieved by the policies~$\mu_1 \in \gP(\gS)$ and~$\mu_2 \in \gP(\gS)$ respectively. Then, $P(\xi^1) = \gJ(\mu_1)$ with~$-H(\pi_b,\mu_1)+\rvkappa_0\geq \xi^1$ and~$P(\xi^2) = \gJ(\mu_2)$ with~$-H(\pi_b,\mu_2)+\rvkappa_0\geq \xi^2$. To establish \Eqref{concave} it suffices to show that for every $t\in(0,1)$ there exists a policy $\mu_t$ such that $-H(\pi_b,\mu_t)+\rvkappa_0\geq t\xi^1+(1-t)\xi^2$ and $\gJ(\mu_t)=t\gJ(\mu_1)+(1-t)\gJ(\mu_2)$. Notice that any policy $\mu_t$ satisfying the previous conditions is a feasible policy for the slack $-\rvkappa_0 +t \xi^1+(1-t)\xi^2$. Hence, by definition of the perturbed function~\Eqref{concave}, it follows that
\begin{equation}
P\left[ t \xi^1 + (1-t) \xi^2 \right] \geq \gJ(\mu_t) =t\gJ(\mu_1)+(1-t)\gJ(\mu_2)=t P\left( \xi^1 \right) + (1-t) P\left( \xi^2 \right).
  \end{equation}
If such a policy exists, the previous equation implies \Eqref{concave}. Thus, to complete the proof of the result we need to establish its existence. To do so we start by formulating the objective $\gJ(\mu)$ as a linear function.

After sufficient training, we can suppose  $Q_\phi(\vs,\va)$ as an unbiased approximation of $Q^\mu(\vs,\va)$, 
\begin{equation}
Q_\phi(\vs,\va)\approx Q^\mu(\vs, \va) = \E_{\tau\sim d_0,\mu,\gT}\sbr{\sum_{j=0}^{J}\gamma^j r(\vs_{j}, \va_{j}\vert \vs_0=\vs,\va_0=\va)},
\end{equation}
where $\tau$ denotes the trajectory generated by policy $\mu(\va\vert\vs)$, start state distribution $d_0$ and environment dynamics $\gT(\vs'\vert\vs,\va)$.

So the objective function \Eqref{obj} can be written as 
\begin{equation}
\label{affine}
    \begin{aligned}
        J(\mu) &= \int_\mathcal{S} d_0(\vs) \int_\mathcal{A}Q^\mu(\vs, \va) d\va d\vs\\
        &= \sum_{t=0}^\infty\gamma^t\E_{\vs_t\sim p_\mu(\vs_t=s)}\sbr{\E_{\va\sim\mu(\va_t\vert\vs_t)}r(\vs_t,\va_t)}\\
&=\sum_{t=0}^{\infty}\gamma^t\sbr{\int_{\gS\times\gA}p_\mu(\vs_t=\vs)\mu(\va_t=\va\vert\vs_t=\vs)r(\vs,\va)d\va d\vs} \\ &=\int_{\gS\times\gA}\sbr{\sum_{t=0}^{\infty}\gamma^tp_\mu(\vs_t=\vs)\mu(\va_t=\va\vert\vs_t=\vs)r(\vs,\va)d\va d\vs} \\
&= \frac{1}{1-\gamma}\E_{\vs,\va\sim\rho^\mu(\vs,\va)}\sbr{r(\vs,\va)}\\
        &= \frac{1}{1-\gamma}\int_{\gS\times\gA}\rho^\mu(\vs,\va)r(\vs,\va)d\va d\vs,
    \end{aligned}
\end{equation}
which implies the objective function \Eqref{obj} equivalent to 
\begin{equation}
    \max_{\rho\in\gR}\frac{1}{1-\gamma}\int_{\gS\times\gA}\rho^\mu(\vs,\va)r(\vs,\va)d\va d\vs,
    \label{rho_obj}
\end{equation}
where $\gR$ is the set of occupation measures, which is convex and compact.[~\citet{borkar1988convex}, Theorem 3.1]

So \Eqref{rho_obj} is a linear function on $\rho(\vs,\va)$. Let $\rho_1(\vs,\va),\rho_2(\vs,\va)\in\gR$ be the occupation measures associated to $ \mu_1$ and $\mu_2$. Since, $\gR$ is convex, there exists a policy $\mu_t$ such that its corresponding occupation measure is $\rho_t(\vs,\va)= t\rho_1(\vs,\va)+(1-t)\rho_2(\vs,\va)\in\gR$ and $\gJ(\mu_t)=t\gJ(\mu_1)+(1-t)\gJ(\mu_2)$. To prove \Eqref{concave}, it suffices to show such $\mu_t$ satisfies $-H(\pi_b,\mu)+\rvkappa_0\geq t\xi^1+(1-t)\xi^2$. Using Assumption \ref{ass2}, we can get 


\begin{equation}
\begin{aligned}
        -H(\pi_b,\mu)&= \int_{\gS\times\gA}\rho^{\pi_b}(\vs,\va)\log\frac{\rho^\mu_t(\vs,\va)}{\rho^{\pi_b}(\vs)}d\va d\vs\\
        &=\int_{\gS\times\gA}\rho^{\pi_b}(\vs,\va)\log\frac{t\rho_1(\vs,\va)+(1-t)\rho_2(\vs,\va)}{\rho^{\pi_b}(\vs)}d\va d\vs\\
        &\underbrace{\geq t\int_{\gS\times\gA}\rho^{\pi_b}(\vs,\va)\log\frac{\rho_1(\vs,\va)}{\rho^{\pi_b}(\vs)}d\va d\vs+}_{\quad \text{Jensen's inequality.}}\\
        &\underbrace{(1-t)\int_{\gS\times\gA}\rho^{\pi_b}(\vs,\va)\log\frac{\rho_2(\vs,\va)}{\rho^{\pi_b}(\vs)}d\va d\vs}_{\quad \text{Jensen's inequality.}}\\
        &\geq t\xi^1+(1-t)\xi^2 - \rvkappa_0\\
\end{aligned}
\end{equation}
This completes the proof that the perturbation function is concave and according to Lemma \ref{T:strongDuality} strong duality for \Eqref{cps} holds.

Finally, we prove \Eqref{cps}, which has the same optimal policy $\mu$ with \Eqref{eq:minmax}.

For \Eqref{eq:minmax}, if strong duality holds and a dual optimal solution $\lambda^*$ exists, then any primal optimal point of \Eqref{cps} is also a maximizer of $\gL(\mu,\lambda^*)$, this is obvious through the KKT conditions.[~\citet{boyd2004convex} Ch.5.5.5.]

So if we can prove that the solution of the $\gL(\mu,\lambda^*)$ is unique, we can compute the optimal policy of \Eqref{cps} from a dual problem \Eqref{eq:minmax}: 
\begin{equation}
\begin{aligned}
    \gL(\mu,\lambda^*) &= J(\mu)+\lambda^*(\rvkappa_0-H(\pi_b,\mu))\\
    &=\frac{1}{1-\gamma}\sbr{\underbrace{\int_{\gS\times\gA}\rho^\mu(\vs,\va)r(\vs,\va)d\va d\vs}_{\quad \text{Using \Eqref{affine}.}}+\lambda^*(\bar{\rvkappa_0}-\int_{\gS\times\gA}\rho^{\pi_b}(\vs,\va)\log\frac{\rho^\mu(\vs,\va)}{\rho^{\pi_b}(\vs)}d\va d\vs)}.\\
\end{aligned}
\label{lagrange}
\end{equation}
According to \Eqref{affine} $J(\mu)$ is a linear function, and negative relative entropy is concave, this implies that $\gL(\mu,\lambda^*)$ is a strictly concave function of $\rho^\mu$ with only a unique maximum value and since we have that the occupancy measure $\rho^\mu(\vs,\va)$ has a one-to-one relationship with policy $\mu$, we can obtain the unique $\mu$ corresponding to $\rho^\mu$. 
\end{proof}

\subsection{Proposition~\ref{proposition1}}
\begin{proof}
\begin{equation}
\begin{aligned}
H(\pi_b,\mu)&=-\E_{\vs\sim \rho^{\pi_b}(\vs),\va\sim\pi_b(\cdot\vert\vs)}\sbr{\log\mu(\cdot\vert\vs)}\\
&=-\E_{\vs\sim \rho^{\pi_b}(\vs)}\sbr{\int \pi_b(\va\vert\vs)d\va\log \int \pi_b(\va^{1:T}\vert\va,\vs)\frac{\mu(\va^{0:T}\vert\vs)}{\pi_b(\va^{1:T}\vert\va,\vs)}d\va^{1:T}}&\\
&\leq \E_{\vs\sim \rho^{\pi_b}(\vs)}\sbr{\underbrace{\int \pi_b(\va^{0:T})\log [\frac{\pi_b(\va^{1:T}\vert\va,\vs)}{\mu(\va^{0:T}\vert\vs)}]d\va^{0:T}}_{\quad \text{Jensen's inequality}}}=K.& \\
\end{aligned}
\label{step1}
\end{equation}
$H(\pi_b,\mu)=K$ is true if and only if $\pi_b(\va^{1:T}\vert\va^0,\vs)=\mu(\va^{1:T}\vert\va^0,\vs)$. By letting the KL divergence between $\pi_b$ and $\mu$ be small enough, we can approximate the $H(\pi_b, \mu)$ with $K$. Furthermore, we can use the variational lower bound $K$ to approximate the KL divergence~\citep{ho2020}:

\begin{align*}
H(\pi_b,\mu)\approx K 
&= \E_{\vs\sim \rho^{\pi_b}(\vs)}\sbr{\mathbb{E}_{\pi_b(\va^{0:T})} \Big[ \log\frac{\pi_b(\va^{1:T}\vert\va,\vs)}{\mu(\va^{0:T}\vert\vs)} \Big]}=\E_{\vs\sim \rho^{\pi_b}(\vs)}\sbr{\mathbb{E}_{\pi_b(\va^{0:T})} \Big[ \log\frac{\pi_b(\va^{1:T}\vert\va^0,\vs)}{\mu(\va^{0:T}\vert\vs)} \Big]} \\
&= \mathbb{E}_{\rho^{\pi_{b}}(\vs,\va)} \Big[ \log\frac{\prod_{t=1}^T \pi_b(\va^t\vert\va^{t-1})}{ \mu(\va^T) \prod_{t=1}^T \mu(\va^{t-1} \vert\va^t) } \Big] \\
&= \mathbb{E}_{\rho^{\pi_{b}}(\vs,\va)} \Big[ -\log \mu(\va^T) + \sum_{t=1}^T \log \frac{\pi_b(\va^t\vert\va^{t-1})}{\mu(\va^{t-1} \vert\va^t)} \Big] \\
&= \mathbb{E}_{\rho^{\pi_{b}}(\vs,\va)} \Big[ -\log \mu(\va^T) + \sum_{t=2}^T \log \frac{\pi_b(\va^t\vert\va^{t-1})}{\mu(\va^{t-1} \vert\va^t)} + \log\frac{\pi_b(\va^1 \vert \va^0)}{\mu(\va^0 \vert \va^1)} \Big] \\
&= \mathbb{E}_{\rho^{\pi_{b}}(\vs,\va)} \Big[ -\log \mu(\va^T) + \sum_{t=2}^T \log \Big( \frac{\pi_b(\va^{t-1} \vert \va^t, \va^0)}{\mu(\va^{t-1} \vert\va^t)}\cdot \frac{\pi_b(\va^t \vert \va^0)}{\pi_b(\va^{t-1}\vert\va^0)} \Big) + \log \frac{\pi_b(\va^1 \vert \va^0)}{\mu(\va^0 \vert \va^1)} \Big] \\
&= \mathbb{E}_{\rho^{\pi_{b}}(\vs,\va)} \Big[ -\log \mu(\va^T) + \sum_{t=2}^T \log \frac{\pi_b(\va^{t-1} \vert \va^t, \va^0)}{\mu(\va^{t-1} \vert\va^t)} + \sum_{t=2}^T \log \frac{\pi_b(\va^t \vert \va^0)}{\pi_b(\va^{t-1} \vert \va^0)} + \log\frac{\pi_b(\va^1 \vert \va^0)}{\mu(\va^0 \vert \va^1)} \Big] \\
&= \mathbb{E}_{\rho^{\pi_{b}}(\vs,\va)} \Big[ -\log \mu(\va^T) + \sum_{t=2}^T \log \frac{\pi_b(\va^{t-1} \vert \va^t, \va^0)}{\mu(\va^{t-1} \vert\va^t)} + \log\frac{\pi_b(\va^T \vert \va^0)}{\pi_b(\va^1 \vert \va^0)} + \log \frac{\pi_b(\va^1 \vert \va^0)}{\mu(\va^0 \vert \va^1)} \Big]\\
&= \mathbb{E}_{\rho^{\pi_{b}}(\vs,\va)} \Big[ \log\frac{\pi_b(\va^T \vert \va^0)}{\mu(\va^T)} + \sum_{t=2}^T \log \frac{\pi_b(\va^{t-1} \vert \va^t, \va^0)}{\mu(\va^{t-1} \vert\va^t)} - \log \mu(\va^0 \vert \va^1) \Big] \\
&= \mathbb{E}_{\rho^{\pi_{b}}(\vs,\va)} [\underbrace{D_\text{KL}(\pi_b(\va^T \vert \va^0) \parallel \mu(\va^T))}_{L_T} + \sum_{t=2}^T \underbrace{D_\text{KL}(\pi_b(\va^{t-1} \vert \va^t, \va^0) \parallel \mu(\va^{t-1} \vert\va^t))}_{L_{t-1}} \underbrace{- \log \mu(\va^0 \vert \va^1)}_{L_0} ].
\end{align*}
So we can approximate the entropy constraint in \Eqref{cps}:
\begin{equation}
    H(\pi_b,\mu)\approx K = L_T+\sum_{t=2}^TL_{t-1}+L_0.
    \label{approx}
\end{equation}
Following DDPM~\citep{ho2020} use the random sample to approximate the $\sum_{t=2}^TL_{t-1}+L_0$, we have
\begin{equation}
\begin{aligned}
&\sum_{t=2}^TL_{t-1}+L_0\approx\gL_c(\pi_b,\mu)\\
&=\E_{i \sim \sbr{1:N}, \rvepsilon \sim \gN(\mathbf{0}, \mI), (\vs, \va) \sim \gD} \sbr{|| \rvepsilon - \rvepsilon_\theta(\sqrt{\bar{\alpha}_i} \va + \sqrt{1 - \bar{\alpha}_i}\rvepsilon, \vs, i) ||^2}.
\end{aligned}
\label{loss2}
\end{equation}
Let $L_T=c$, combining \Eqref{approx} and \Eqref{loss2} we can get
\begin{equation}
    H(\pi_b,\mu)\approx c+\gL_c(\pi_b,\mu).
\end{equation}
\end{proof}
\subsection{Theorem~\ref{T:convergence}}
The proof of  Theorem \ref{T:convergence} is based on \citet{paternain2019constrained}[Appendix A].

We first claim lemma~\ref{lemma_meassure_bound} to show the error introduced by the parametrization is bound by a constant.
\begin{lemma}\label{lemma_meassure_bound}
  Let $\rho$ and $\rho_\theta$ be occupation measures induced by the policies $\pi\in\gP(\gS)$ and $\pi_{\theta}$ respectively, where $\pi_{\theta}$ is an $\epsilon$- parametrization of $\pi$. Then, it follows that
  \begin{equation}\label{eqn_occupation_bound}
\int_{\gS\times\gA} |\rho(s,a)-\rho_\theta(s,a)| \,dsda \leq \frac{\epsilon}{1-\gamma}.
    \end{equation}
\end{lemma}
See proof in \citet{paternain2019constrained}[Appendix A].

We next introduce Theorem \ref{T:parametrization}, which shows that the dual gap introduced after parameterization is bounded by a linear function of the model's expressive capability. 
\begin{theorem}\label{T:parametrization}
  Suppose that~$r$ is bounded  by constants~$R>0$ and $B = \max_{\vs,\va}\rho^{\pi_b}(\vs,\va)1/\rho(\vs,\va)$. Let $\lambda^\star$ be the solution to the dual problem associated with \Eqref{cps}.
  Then, under the hypothesis of Theorem \ref{theorem2} it follows that 
  \begin{equation}\label{eqn_azdg}
    P^\star\geq D_\theta^\star \geq P^\star - \left(R+\left\|{\lambda^\star}\right\|_1B\right)\frac{\epsilon}{1-\gamma},
  \end{equation}
  where~$P^\star$ is the optimal value of~\Eqref{cps}, and~$D_{\theta}^\star$ the value of the parametrized dual problem \Eqref{eta2}.
  \end{theorem}
Note that Theorem \ref{T:parametrization} is actually based on Theorem 2 of \citet{paternain2019constrained}, and the proof is as follows.
 
\begin{proof}[Proof of Theorem~\ref{T:parametrization}]

  Notice that the dual functions $d(\lambda)$ and $d_\theta(\lambda)$ associated to the problems \Eqref{cps} and \Eqref{mu2} respectively are such that for every $\lambda$ we have that
  $  d_\theta(\lambda) \leq d(\lambda)$. The latter follows from the fact that the set of maximizers of the Lagrangian for the parametrized policies is contained in the set of maximizers of the non-parametrized policies. In particular, this holds for $\lambda^\star$ the solution of the dual problem associated to \Eqref{cps}. Hence we have the following sequence of inequalities
  \begin{equation}
D^\star = d(\lambda^\star) \geq d_{\theta}(\lambda^\star) \geq D^\star_\theta, 
    \end{equation}
  where the last inequality follows from the fact that $D^\star_\theta$ is the minimum of \Eqref{eta2}. The zero duality gap established in Theorem \ref{theorem2} completes the proof of the upper bound for $D_\theta^\star$. We next work towards proving the lower bound for $D_\theta^\star$. Let us next write the dual function of the parametrized problem \Eqref{mu2} as
  \begin{equation}
 d_{\theta}(\lambda) =  d(\lambda)-\left(\max_{\mu\in\gP(\gS)} \gL(\mu,\lambda) - \max_{\theta}\gL_\theta(\theta,\lambda)\right) 
\end{equation}
  Let $\mu^\star\triangleq \argmax_{\mu\in\gP(\gS)} \gL(\mu,\lambda)$ and let $\theta^\star$ be an $\epsilon$-approximation of $\mu^\star$. Then, by definition of the maximum it follows that 
    \begin{equation}\label{eqn_relationship_duals}
 d_{\theta}(\lambda) \geq  d(\lambda)-\left( \gL(\mu^\star,\lambda) - \gL_\theta(\theta^\star,\lambda)\right) 
    \end{equation}
We next work towards a bound for $\gL(\mu^\star,\lambda)-\gL_\theta(\theta^\star,\lambda)$. To do so, notice that we can write the difference in terms of the occupation measures where $\rho^\star$ and $\rho_{\theta}^\star$ are the occupation measures associated to the the policies $\mu^\star$ and the policy $\mu_{\theta^\star}$ 
\begin{equation}
\gL(\mu^\star,\lambda)-\gL_\theta(\theta^\star,\lambda) = \int_{\gS\times\gA} r\left( d\rho^\star(\lambda) -d\rho_{\theta}^\star(\lambda)\right) +\lambda\rho^{\pi_b}(\vs,\va)d(\log\rho^\star(\lambda)-\log\rho_{\theta}^\star(\lambda)).
\end{equation}
Since $\mu_{\theta^\star}$ is by definition an $\epsilon$ approximation of $\mu^\star$ it follows from Lemma \ref{lemma_meassure_bound} that
    \begin{equation}
      \int_{\gS\times\gA} \left|d \rho^\star(\lambda)-d\rho_\theta^\star(\lambda)\right|  \leq \frac{\epsilon}{1-\gamma}.
    \end{equation}
Using the bounds on the reward functions and occupation measures we can upper bound the difference $\gL(\pi^\star,\lambda)-\gL_\theta(\theta^\star,\lambda)$ by
\begin{equation}
  \gL(\mu^\star,\lambda)-\gL_\theta(\theta^\star,\lambda) \leq \left(R+\left\|\lambda\right\|_1B\right)\frac{\epsilon}{1-\gamma}.
\end{equation}
The second term follows from $(a-b)1/a\leq\log a -\log b \leq (a-b)1/b, a\geq b$.
Combining the previous bound with \Eqref{eqn_relationship_duals} we can lower bound $d_\theta(\lambda)$ as 
  \begin{equation}\label{eqn_aux_proof_almost_not_dualitygap}
d_\theta(\lambda)\geq d(\lambda) - \left(R+\left\|\lambda\right\|_1B\right)\frac{\epsilon}{1-\gamma}
    \end{equation}

Since the previous expression holds for every $\lambda$, in particular it holds for $\lambda_\theta^\star$, the dual solution of the parametrized problem \Eqref{eta2}. Thus, we have that 
  \begin{equation}\label{eqn_aux_proof_almost_not_dualitygap}
D_\theta^\star\geq d(\lambda) - \left(R+\left\|\lambda\right\|_1B\right)\frac{\epsilon}{1-\gamma}
    \end{equation}
    Recall that $\lambda^\star =\argmin d(\lambda)$, and use the definition of the dual function to lower bound $D_{\theta}^\star$ by 
    \begin{equation}
      D_{\theta}^\star\geq \max_{\mu\in\gP(\gS)} \gJ(\mu) + \lambda^\star\left(\rvkappa_0-H(\pi_b,\mu)\right) - \left(R+\left\|\lambda\right\|_1B\right)\frac{\epsilon}{1-\gamma}. 
    \end{equation}
    By definition of maximum, we can lower bound the previous expression by substituting by any $\mu \in\gP(\gS)$. In particular, we select $\mu^\star$ the solution to \Eqref{cps}
        \begin{equation}
      D_{\theta}^\star\geq \max_{\mu\in\gP(\gS)} \gJ(\mu^\star) + \lambda^\star\left(\rvkappa_0-H(\pi_b,\mu^\star)\right) - \left(R+\left\|\lambda\right\|_1B\right)\frac{\epsilon}{1-\gamma}. 
    \end{equation}
        Since $\mu^\star$ is the optimal solution to \Eqref{cps} it follows that $\rvkappa_0-H(\pi_b,\mu)\geq 0$ and since $\lambda^\star\geq 0$ the previous expression reduces to 

      \begin{equation}
      D_{\theta}^\star\geq \gJ(\mu^\star) - \left(R+\left\|\lambda\right\|_1B\right)\frac{\epsilon}{1-\gamma} = P^\star-\left(R+\left\|\lambda\right\|_1B\right)\frac{\epsilon}{1-\gamma}
    \end{equation}
  Which completes the proof of the result
\end{proof}

%
Next, we claim that the subgradient calculation error in the dual update (line $14$ in Algorithm \ref{alg:main}) through Proposition \ref{prop_subgradient}
\begin{assumption}\label{assumption_neural_network}
Let $\mu_\theta$ be a parametrization of functions in $\gP(\gS)$ and let $\gL_\theta(\theta,\lambda)$ with $\lambda\in\mathbb{R}^m_+$ be the Lagrangian associated to \Eqref{mu2}. Denote by $\theta^\star(\lambda),\theta^\dagger(\lambda)\in\mathbb{R}^P$ the maximum of $\gL(\theta,\lambda)$ and a local maximum respectively achieved by a generic reinforcement learning algorithm. Then, there exists $\delta>0$ such that for all $\lambda\in\mathbb{R}^m_+$ it holds that $\gL_\theta(\theta^\star(\lambda),\lambda) \leq \gL_\theta(\theta^\dagger(\lambda),\lambda)+\delta$.
\end{assumption}
\begin{proposition}\label{prop_subgradient}
  Under Assumption \ref{assumption_neural_network}, the constraint in \Eqref{mu2} evaluated at a local maximizer of Lagrangian $\theta^\dagger(\lambda)$ approximate the subgradient of the dual function \Eqref{eta2}. In particular, it follows that
  \begin{equation}
    d_\theta(\lambda) - d_\theta(\lambda_\theta^\star) \leq  \left(\lambda-\lambda_\theta^\star\right)\left( \rvkappa_0-H(\pi_b,\mu)\right)+\delta.
  \end{equation}
  \end{proposition}
See proof in \citet{paternain2019constrained} (Appendix A).

Finally, we prove Theorem \ref{T:convergence}.
\begin{proof}[Proof of Theorem \ref{T:convergence}]
  We start by showing the lower bound, which in fact holds for any $\lambda$. Notice that for any $\lambda$ and by definition of the dual problem it follows that $d_\theta(\lambda)\geq D^\star_\theta$. Combining this bound with the result of Theorem \ref{T:parametrization} it follows that
        \begin{equation}
          d_\theta(\lambda) \geq P^\star - \left(R+\left\|{\lambda^\star}\right\|_1B\right)\frac{\epsilon}{1-\gamma}.
\end{equation}
        To show the upper bound we start by writing the difference between the dual multiplier $k+1$ and the solution of \Eqref{eta2} in terms of the iteration at time $k$. Since $\lambda_{\theta}^\star\in\mathbb{R}^+$ and using the non-expansive property of the projection it follows that 
  \begin{equation}
\left\|\lambda_{k+1}-\lambda^\star_\theta\right\|^2 \leq \left\|\lambda_{k}-\eta \left(\rvkappa_0-H(\pi_b,\theta^\dagger(\lambda_k))\right)-\lambda^\star_\theta\right\|^2 
  \end{equation}
  Expanding the square and using that $C =\left( \rvkappa_0+\int_{\gS\times\gA}\rho^{\pi_b}(\vs,\va)\log\pi_b(\va\vert\vs)d\va d\vs\right)^2$ is a bound which is calculated by the KL divergence in \Eqref{kl} is nonnegative on the norm squared of $\rvkappa_0-H(\pi_b,\theta^\dagger(\lambda_k))$ it follows that
  \begin{equation}
\left\|\lambda_{k+1}-\lambda^\star_\theta\right\|^2 \leq    \left\|\lambda_k-\lambda_\theta^\star\right\|^2 - 2\eta\left(\lambda_k-\lambda_\theta^\star\right) \left(\rvkappa_0-H(\pi_b,\theta^\dagger(\lambda_k))\right)+\eta^2 C. 
    \end{equation}
  Using the result of Proposition \ref{prop_subgradient} we can further upper bound the inner product in the previous expression by the difference of the dual function evaluated at $\lambda_k$ and $\lambda_\theta^\star$ plus $\delta$, the error in the solution of the primal maximization,
    \begin{equation}
\left\|\lambda_{k+1}-\lambda^\star_\theta\right\|^2 \leq    \left\|\lambda_k-\lambda_\theta^\star\right\|^2 +2\eta\left(\delta+d_\theta(\lambda_{\theta}^\star)-d_\theta(\lambda_k)\right)+\eta^2 C. 
    \end{equation}
    Defining $\alpha_k = 2(\delta+d_\theta(\lambda_{\theta}^\star)-d_\theta(\lambda_k))+\eta C$ and writing recursively the previous expression yields 
        \begin{equation}\label{eqn_dual_reduction}
\left\|\lambda_{k+1}-\lambda^\star_\theta\right\|^2 \leq    \left\|\lambda_0-\lambda_\theta^\star\right\|^2 +\eta \sum_{j=0}^k\alpha_j.
        \end{equation}
        Since $d_\theta(\lambda_\theta^\star)$ is the minimum of the dual function, the difference $d_\theta(\lambda_{\theta}^\star)-d_\theta(\lambda_k)$ is always negative. Thus, when $\lambda_k$ is not close to the solution of the dual problem $\alpha_k$ is negative. The latter implies that the distance between $\lambda_k$ and $\lambda_\theta^\star$ is reduced by virtue of \Eqref{eqn_dual_reduction}. To be formal, for any $\varepsilon>0$, when $a_j>-2\varepsilon$ we have that 
          \begin{equation}
d_\theta(\lambda_j) - d_\theta(\lambda_\theta^\star) \leq \eta\frac{C}{2}+\delta+\varepsilon.
            \end{equation}
          Using the result of Theorem \ref{T:parametrization} we can upper bound $D_\theta^\star$ by $P^\star$ which establishes the neighborhood defined in \Eqref{eqn_convergence_neighborhood}. We are left to show that the number of iterations required to do so is bounded by
                  \begin{equation}
          K \leq \frac{\left\|\lambda_0-\lambda_\theta^\star\right\|^2}{2\eta \varepsilon}.
        \end{equation}
To do so, let $K>0$ be the first iterate in the neighborhood \Eqref{eqn_convergence_neighborhood}. Formally, $K = \min_{j\in\mathbb{N}} \alpha_j>-2\varepsilon$. Then it follows from the recursion that
        \begin{equation}
\left\|\lambda_{K}-\lambda^\star_\theta\right\|^2 \leq    \left\|\lambda_0-\lambda_\theta^\star\right\|^2 -2K\eta \varepsilon.
        \end{equation}
Since $\left\|\lambda_K-\lambda_\theta^\star\right\|^2$ is positive the previous expression reduces to $2K\eta\epsilon \leq \left\|\lambda_0-\lambda_\theta^\star\right\|^2$. Which completes the proof of the result.  
\end{proof}


\section{Notation Table}
\label{notation}
\begin{table}[htbp]
\begin{center}
\caption{Table of Notation in paper}
\begin{tabular}{r c p{10cm} }
\toprule
$ H(\pi_b,\mu)$ & $\triangleq$ & $-\E_{\vs\sim \rho^{\pi_b}(\vs),\va\sim\pi_b(\cdot\vert\vs)}\sbr{\log\mu(\cdot\vert\vs)}$.\\
$\mu$, $\mu(\va\vert\vs)$ or $\mu(\cdot\vert\vs)$ & $\triangleq$ & policy. \\
$\pi_b$, $\pi_b(\va\vert\vs)$ or $\pi_b(\cdot\vert\vs)$ & $\triangleq$ & behavior policy. \\
$p_\mu(\vs_t=\vs)$ & $\triangleq$ & Probability of landing in state $\vs$ at time $t$, when following policy $\mu$ from an initial state sampled from $d_0$, in an environment with transition dynamics $\gT$.\\
$\gT$, $\gP(\vs'\vert\vs,\va)$ or $p(\vs'\vert\vs,\va)$ & $\triangleq$ & transition dynamics.\\  
$d_0(\vs)$ & $\triangleq$ & initial state distribution of behavior policy.\\
$\rho^\mu(\vs)$ & $\triangleq$ & $\rho^\mu(\vs)=\sum_{t=0}^\infty\gamma^t p_\mu(\vs_t =\vs)$ is the unnormalized discounted state visitation frequencies.\\
$\rho^\mu(\vs,\va)$ & $\triangleq$ & $\rho^\mu(\vs,\va)=(1-\gamma)\sum_{t=0}^\infty\gamma^t p_\mu(\vs_t =\vs)\mu(\va\vert\vs)=(1-\gamma)\rho^\mu(\vs)\mu(\va\vert\vs)$ is the occupation measure of policy $\mu$.\\
$Q_\phi(\vs,\va)$ & $\triangleq$ & parameterized state action function. \\
$Q_{\phi^{'}}(\vs,\va)$ & $\triangleq$ & parameterized target state action function. \\
$\mu_\theta(\va\vert\vs)$ & $\triangleq$ & parameterized policy. \\
$\mu_{\theta^{'}}(\va\vert\vs)$ & $\triangleq$ & parameterized target policy. \\
$\rvepsilon_\theta(\vx_i, i) $ & $\triangleq$ & parameterized Gaussian noise predict network in diffusion model. \\
$f_\phi(\vy\vert\vx_i)$ & $\triangleq$ & parameterized classifier. \\
\bottomrule
\end{tabular}
\end{center}
\label{tab:TableOfNotationForMyResearch}
\end{table}
\section{EXPERIMENTAL DETAILS}
\label{details}
We follow the~\citet{wang2022b} to build our policy with an MLP-based conditional diffusion model. Following the DDPM, we recover the action from a residual network $\rvepsilon(\va^i,\vs, i)$, and we model the $\rvepsilon_\theta$ as a 3-layers MLPs with Mish activations. All the hidden units have been set to 256. We also follow the~\citet{fujimoto2018} 
to build two Q networks with the same MLP setting as our critic network. All the networks are optimized through Adam~\citep{kingma2014adam}. We provide all the hyperparameters in Table \ref{tab:hyperparameter}.

We train for $1000$ epochs ($2000$ for Gym tasks). Each epoch consists of $1000$ gradient steps with batch size $256$. The training in MuJoCo locomotion is usually quite stable as shown in Figure~\ref{ablation_T}. However, the training for AntMaze is relatively unstable due to its sparse reward setting and suboptimal trajectories in the offline datasets.

\begin{table*}[!t]
\centering
\small
        \caption{\small Hyperparameter settings of all selected tasks. Reward Tune with CQL means that we use the reward tuning method in ~\citet{kumar2020}. We also use max Q backup to improve performance in AntMaze tasks. In offline reinforcement learning, it's common to fine-tune rewards for the AntMaze tasks. Additionally, we have observed that some other diffusion-based methods, such as SfBC, perform more reward engineering compared to our method. $\lambda_\text{clip}$ in our paper means that $\lambda \geq \lambda_\text{clip}$.}
\resizebox{1.0\textwidth}{!}{%
\begin{tabular}{llcccccc}
\toprule
\multicolumn{1}{c}{\bf Dataset} & \multicolumn{1}{c}{\bf Environment} & \multicolumn{1}{c}{\bf Learning Rate}& \multicolumn{1}{c}{$\bf{\rvkappa}$}  & \multicolumn{1}{c}{\bf Reward Tune} & \multicolumn{1}{c}{\bf max Q backup} &\multicolumn{1}{c}{\bf Policy evaluation interval} & \multicolumn{1}{c}{$\bf{\lambda_\text{clip}}$} \\
\midrule
Medium-Expert & HalfCheetah    & 3e-4               &  $0.04$ &  None  & False  & $2$  & $0$                  \\
Medium-Expert & Hopper         & 3e-4              &  $0.03$    & None & False & $2$ &$0$          \\
Medium-Expert & Walker2d       & 3e-4         & $0.04$      & None& False& $2$ &$0$            \\
\midrule
Medium        & HalfCheetah    &  3e-4              &  $0.06$     &  None & False   & $2$     & $0$             \\
Medium        & Hopper         &  3e-4             &  $0.05$     &  None     & False   & $2$    & $0$     \\
Medium        & Walker2d        & 3e-4              &  $0.03$     &  None    & False    & $2$    & $0$ \\
\midrule
Medium-Replay & HalfCheetah    &  3e-4        &  $0.06$     &  None    &   False   & $2$   &$0$ \\
Medium-Replay & Hopper         &  3e-4              &  $0.03$     &  None     &   False   & $2$   & $0$      \\
Medium-Replay & Walker2d       &  3e-4             &  $0.03$     &  None     &   False  & $2$    & $0$      \\
\midrule

Default       & AntMaze-umaze  &  3e-4              &  $0.2$     & CQL      & False & $2$  & $0.3$          \\
Diverse       & AntMaze-umaze  &  3e-4        &  $0.09$   & CQL       & True  & $2$ & $0.3$        \\
\midrule  
Play          & AntMaze-medium &  1e-3              &  $0.3$      & CQL      & True  & $2$ & $0.3$           \\
Diverse       & AntMaze-medium &  3e-4              &  $0.2$      & CQL       & True  & $2$ & $0.3$          \\
\midrule
Play          & AntMaze-large  &  3e-4             &  $0.2$      & CQL       & True    & $4$    & $0.3$           \\
Diverse       & AntMaze-large  &  3e-4              &  $0.2$      & CQL       & True   & $4$    & $0.3$           \\

\specialrule{.05em}{.4ex}{.1ex}
\specialrule{.05em}{.1ex}{.65ex}
\end{tabular}
}

        \label{tab:hyperparameter}
\end{table*}

\textbf{Runtime.} We test the runtime of DiffCPS on a RTX 3050 GPU. For algorithm training, the runtime cost of training the gym Locomotion tasks is about  4h26min for $2000$ epochs (2e6 gradient steps), see Table~\ref{table:time} for details.

\begin{table*}[!htbp]
\caption{Runtime of different diffusion-based offline RL methods.}
\centering
\small
\scalebox{0.8}{
\begin{tabular}{lccccccc}
\toprule
\multicolumn{1}{c}{\bf D4RL Tasks}  & \multicolumn{1}{c}{\bf DiffCPS (T=5) }
& \multicolumn{1}{c}{\bf DiffusionQL (T=5) }
& \multicolumn{1}{c}{\bf SfBC (T=5)}   \\
\midrule
\multicolumn{1}{c}{\bf Locomotion Runtime ($1$ epoch)}  &      $7.68$s       &  $5.1$s   &  $8.42$s     \\

\midrule
\multicolumn{1}{c}{\bf AntMaze Runtime ($1$ epoch)} &      $9.96$s       &  $10.5$s &    $10.53$s     \\

\bottomrule
\end{tabular}
}
\label{table:time}
\end{table*}

From Table~\ref{table:time}, it's evident that the runtime of DiffCPS is comparable to other diffusion model-based methods. Further optimization like using Jax and incorporating other diffusion model acceleration tricks can improve the runtime of DiffCPS.

\section{More Toy experiments}
\label{toy_append}
In this chapter, we provide further detailed information about the toy experiments. The noisy circle is composed of a unit circle with added Gaussian noise $\gN(0,0.05)$. The offline data $\gD = \cbr{(\va_i)}_{i=1}^{5000}$ are collected from the noisy circle. We train all methods with 20,000 steps to ensure convergence. The network framework of SfBC remains consistent with the original SfBC paper. Both DQL and DiffCPS use the same MLP architecture.
The code for DQL and SfBC is sourced from the official code provided by the authors, while the code for AWR is adopted from a PyTorch implementation available on GitHub.

\begin{figure}[!t]
\centering
\includegraphics[width = .6\linewidth]{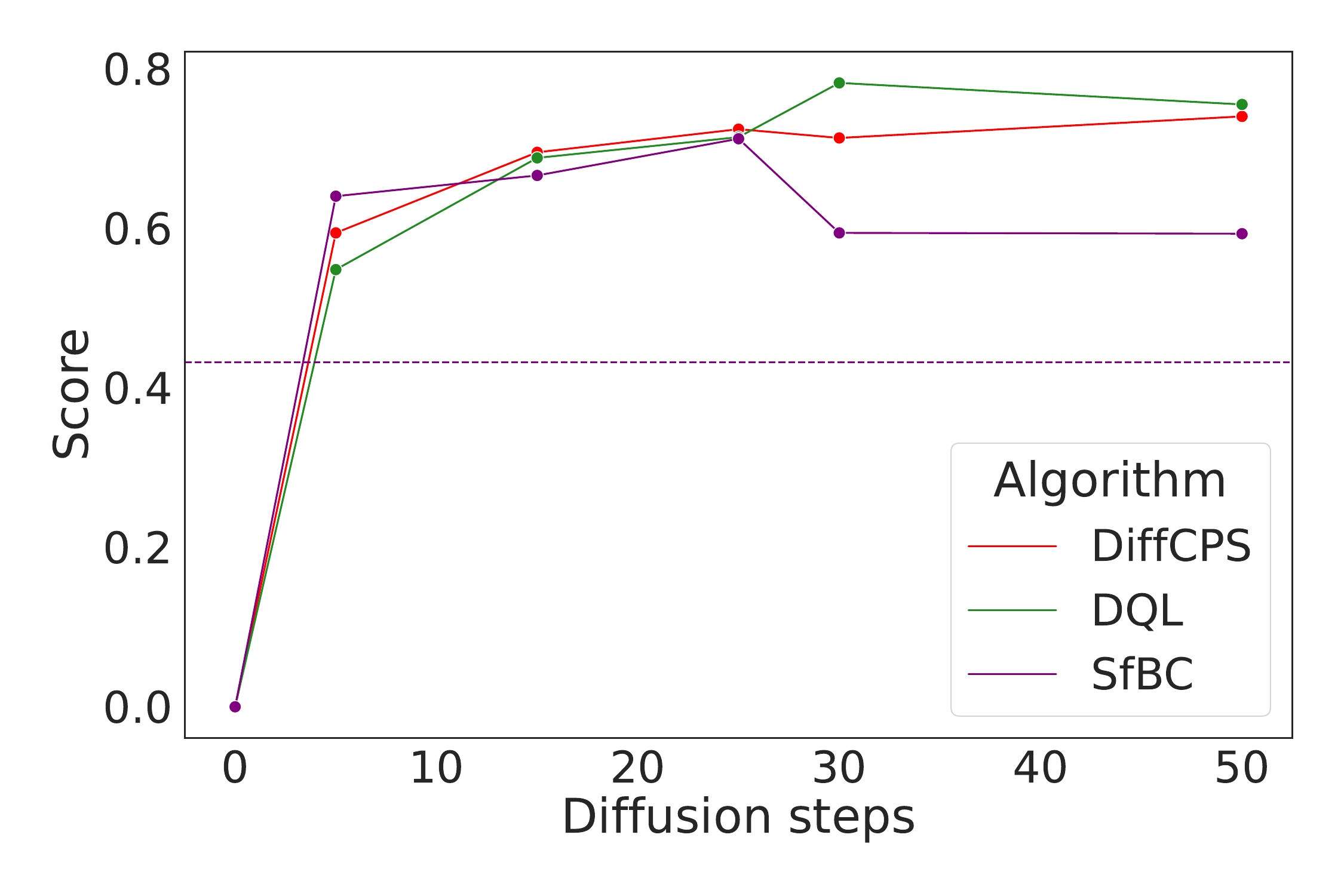}\\
\vspace{-4.mm}
\caption{Evaluation performance of DiffCPS and other baselines on toy bandit experiments. The dashed line represents the score of AWR. We also observe that as $T$ increases, diffusion-based algorithms all experience a certain degree of performance decline, especially SfBC. The reason could be that as $T$ increases, the increased model capacity leads to overfitting the data in the dataset. In the case of SfBC, the presence of sampling errors exacerbates this phenomenon.}
\label{fig:toy_t}
\end{figure}
\begin{figure}[!t]
        \centering
        \includegraphics[width=.9\textwidth]{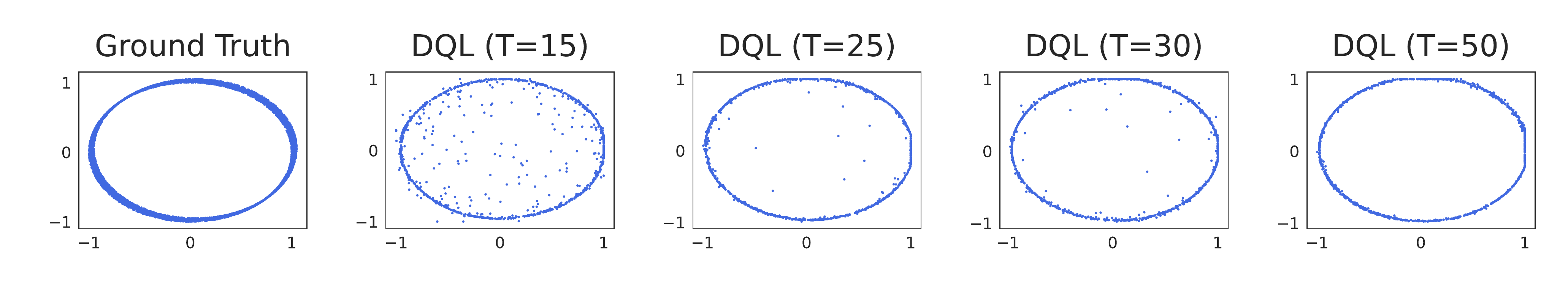} \\
        \includegraphics[width=.9\textwidth]{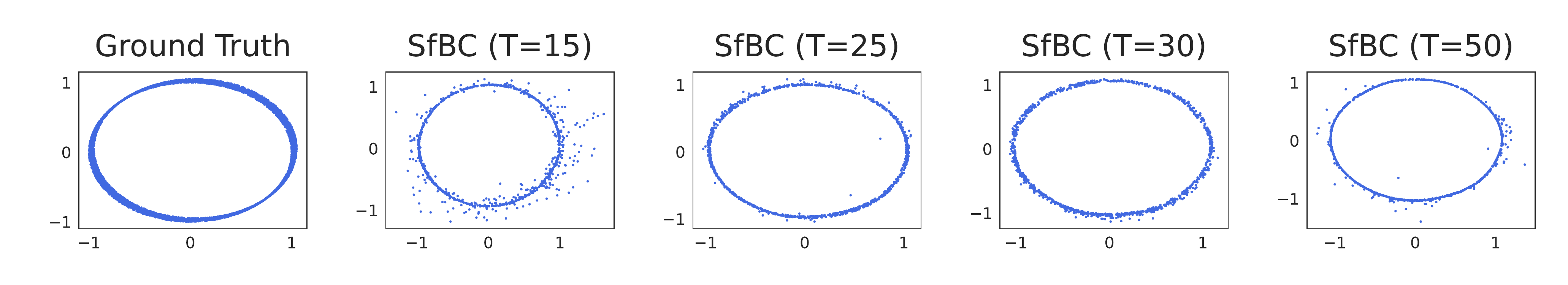}\\
        \includegraphics[width=.9\textwidth]{imgs/diffcps_res.pdf}
        \vspace{-4.mm}
        \caption{Evaluation effect of different diffusion steps on diffusion-based algorithms.
        }
        \label{fig:toy_ablation}
    \end{figure}

In Figure~\ref{fig:toy_ablation}, we put the full results of diffusion-based methods and Figure~\ref{fig:toy_t} displays the scores of diffusion-based algorithms under different diffusion steps, denoted as $T$. The score calculation involves a modified Jaccard Index, which computes how many actions fall into the offline dataset (the minimum distance is less than 1e-8). The dashed line represents the score of AWR.

    \begin{figure}[!t]
        \centering
        \vspace{-4.mm}
        \includegraphics[width=.9\textwidth]{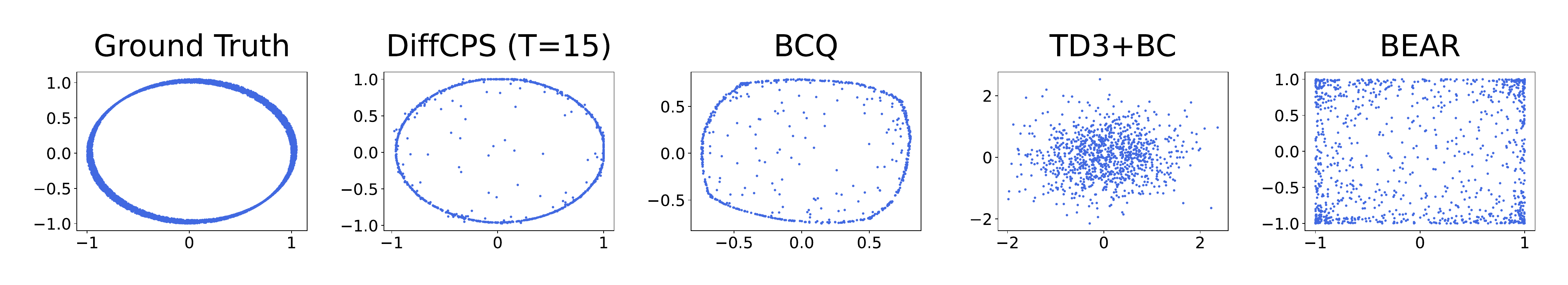} \\
        \vspace{-4.mm}
        \caption{Evaluation effect of prior offline methods.
        }
        \label{fig:toy_ablation2}
\end{figure}
\begin{figure}[!t]
        \centering
        \includegraphics[width=.9\textwidth]{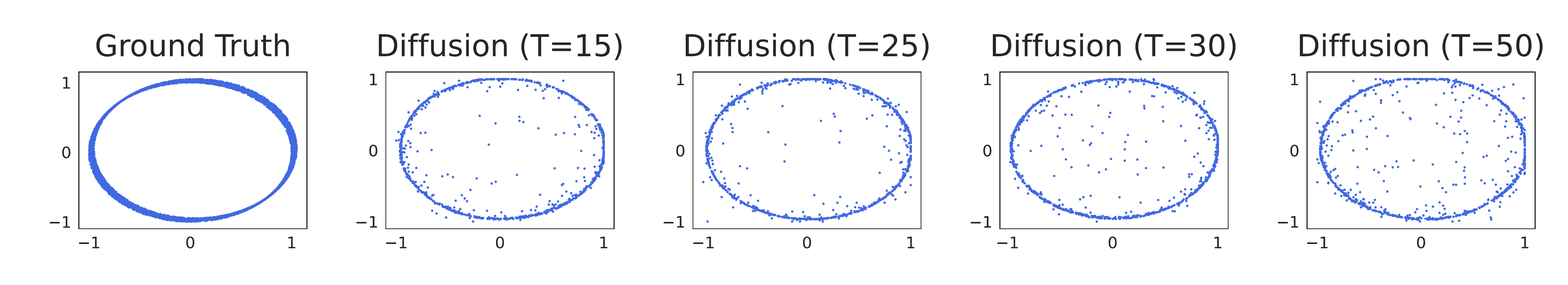}
        \caption{Results of diffusion behavior clone.
        }
        \label{fig:toy_ablation3}
\end{figure}

In Figure~\ref{fig:toy_ablation2}, we compare DiffCPS with prior offline RL methods, including the TD3+BC~\citep{fujimoto2021} with Gaussian policy, BCQ~\citep{fujimoto2019}, and BEAR~\citep{kumar2019stabilizing}. Note that both BCQ and BEAR use the VAE to model behavior policy, and then get the policy by regularizing the cloned behavior policy. However, results in Figure~\ref{fig:toy_ablation2} show that VAE cannot exactly model the multi-modal data in offline datasets. TD3+BC with Gaussian policy also fails to model the optimal behavior due to the limited policy expressivity described in Section~\ref{toy}. 

In Figure~\ref{fig:toy_ablation3}, we show the results of diffusion-based behavior clone. The results in Figure~\ref{fig:toy_ablation3} and Figure~\ref{fig:toy_ablation2} show that the diffusion model can model multi-modal distribution while other methods struggle to capture the multi-modal behavior policy.

\section{Extra Related Work}
\label{extra_related}
\textbf{Offline Model-based RL.} Model-based RL methods represent another approach to address offline reinforcement learning. Similar to model-free offline RL, model-based RL also needs to address extrapolation error.  \citet{kidambi2020morel,yu2020mopo} perform pessimistic planning in the learned model, which penalizes the reward for uncertainty. BREMEN~\citep{matsushima100deployment} uses trust region constraint to update policy in an ensemble of dynamic models to avoid the extrapolation error. VL-LCB~\citep{rashidinejad2021bridging} employs offline value iteration with lower confidence bound to address the extrapolation error. SGP~\citep{suh2023fighting} uses the score function of the diffusion model to approximate the gradient of the proposed uncertainty estimation. Our method differs from them for DiffCPS tackles the diffusion-based constrained policy search problem in offline RL through Lagrange dual and recursive optimization.

\section{Extra Experiments}
\label{extra experiments}
\subsection{Trajectory Stitching}
\begin{figure}[htbp]	
\centering
	\subfigure[Offline Dataset.] 
	{
		\begin{minipage}{6cm}
			\centering          
			\includegraphics[scale=0.4]{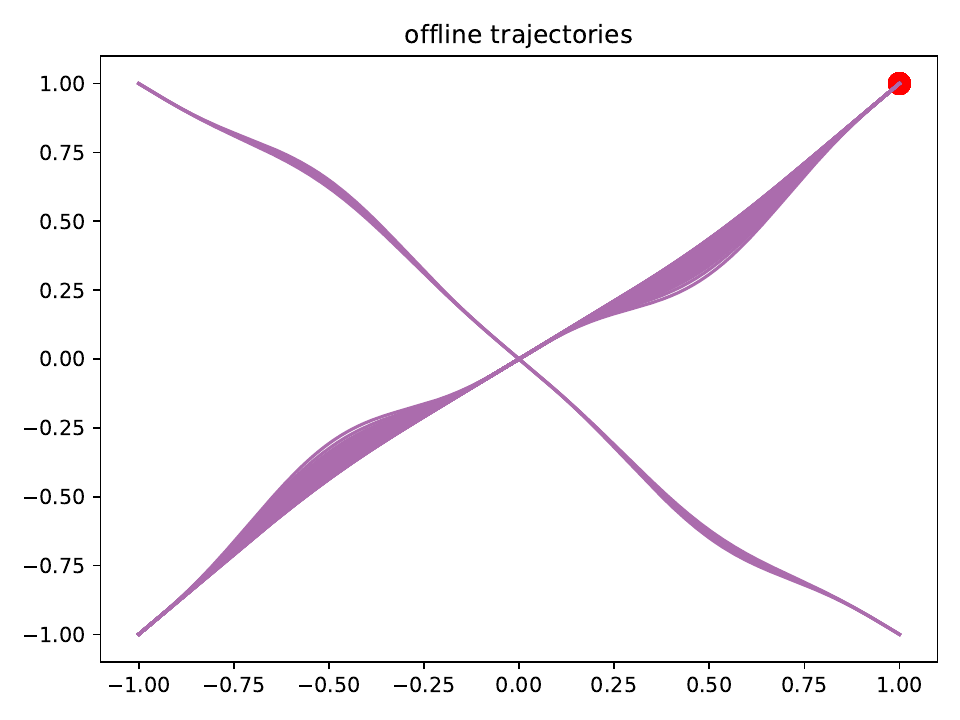}   
		\end{minipage}
	}
	\subfigure[Trajectory of DiffCPS.] 
	{
		\begin{minipage}{7cm}
			\centering      
			\includegraphics[scale=0.4]{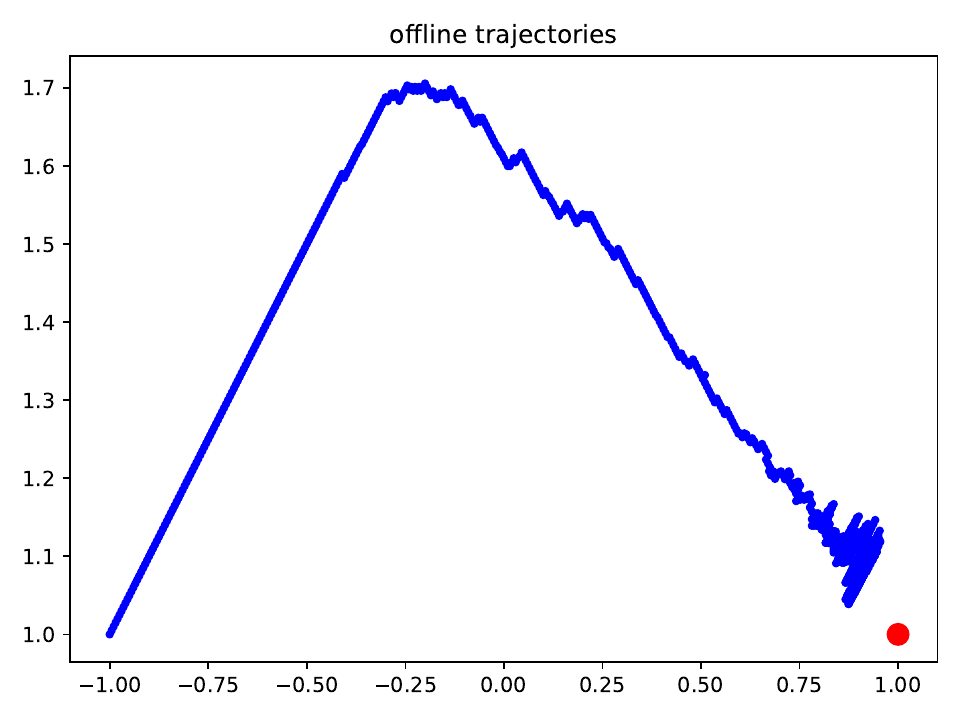}   
		\end{minipage}
	}
	\caption{Trajectory Stitching} %
	\label{fig:10}  
\end{figure}

The offline dataset is depicted in the Figure~\ref{fig:10} (a). 
The reward function is the sum of the negative Euclidean distance from the point $(1,1)$ and a constant, with both actions and states represented as 2D coordinates. The dataset consists only of trajectories forming an X shape. During the evaluation, the starting point is at $(-1,1)$. To achieve maximum reward, the policy needs to stitch together offline data to reach the goal. The trajectories generated by our policy are shown in Figure~\ref{fig:10} (b). We note that DiffCPS successfully combines trajectories from the dataset to reach the vicinity of the target point $(1, 1)$ while avoiding low-reward areas.
\subsection{TriFinger Dataset Experiment} 
The TriFinger dataset~\citep{gurtler_benchmarking_2023} is an offline dataset for the TriFinger robotic arm that includes both real and simulated data. The dataset primarily focuses on two tasks: Push and Lift. We utilize the Push-Sim-Expert dataset to train our DiffCPS, and the network architecture is consistent with the details provided in Appendix~\ref{details}. The goal of the push task is to move the cube to a target location. This task does not require the agent to align the orientation of the cube; the reward is based only on the desired and the achieved position. 

When trained with 2e6 training steps (gradient steps) on the sim-expert dataset, DiffCPS ($T=45$, target\_kl $=0.01$, $\lambda_\text{min}=0$) achieves a success rate of $0.906\pm0.001$ in the push task based on 100 evaluation episodes and 5 random seeds, as detailed in the Table~\ref{table:trifinger}.

\begin{table*}[!htbp]
\centering
\small
\scalebox{0.8}{
\begin{tabular}{lccccccc}
\toprule
\multicolumn{1}{c}{\bf TriFinger-Push}  & \multicolumn{1}{c}{\bf BC }
& \multicolumn{1}{c}{\bf CRR }& \multicolumn{1}{c}{\bf IQL} 
& \multicolumn{1}{c}{\bf DiffCPS (Ours)}   \\
\midrule
\multicolumn{1}{c}{\bf Sim-Expert}  &      $0.83\pm0.02$      &  $\bf{0.94\pm0.04}$   &  $0.88\pm0.04$  &      $\bf{0.906\pm0.001}$   \\

\bottomrule
\end{tabular}
}
\caption{The success rate of DiffCPS and other baselines on TriFinger Push task. Results for other baselines are sourced from \citet{gurtler_benchmarking_2023} and have been carefully tuned. It's worth noting that our DiffCPS still utilizes a simple MLP to represent the policy, suggesting potential benefits from more powerful network architectures. } 
\label{table:trifinger}
\end{table*}
\section{Augmented Lagrangian Method}
\label{alm}
In Algorithm~\ref{alg:main}, we utilize alternating optimization (dual ascent) to solve the dual problem in \Eqref{eq:minmax}. In practice, we can employ the Augmented Lagrangian Method (ALM) to enhance the stability of the algorithm. ALM introduces a strongly convex term (penalty function) into the Lagrangian function to improve convergence. Our experimental results suggest that ALM can enhance training stability but may lead to a performance decrease. This could be attributed to (i) the introduction of the penalty function altering the optimal KL divergence constraint value and (ii) the penalty function making the policy too conservative, thereby preventing the adoption of high-reward actions. We believe that further research will benefit our algorithm by incorporating ALM and other optimization algorithms.

\end{document}